\documentclass[10pt,twocolumn,letterpaper]{article}

 \usepackage[pagenumbers]{cvpr} 
\usepackage{times}
\usepackage{epsfig}
\usepackage{array}     
\usepackage{booktabs}  
\usepackage{multirow}  
\usepackage{makecell}  
\usepackage{graphicx}  
\usepackage{amsmath,amssymb}
\usepackage{algorithm}
\usepackage{algpseudocode}  
\usepackage{enumitem} 
\usepackage{subcaption} 
\usepackage{bm}
\usepackage{placeins}
\usepackage{amsthm}

\newtheorem{Proposition}{Proposition}
\usepackage{booktabs}
\usepackage[numbers,sort&compress]{natbib}
\definecolor{cvprblue}{rgb}{0.21,0.49,0.74}
\usepackage[pagebackref,breaklinks,colorlinks,allcolors=cvprblue]{hyperref}

\def\paperID{14613} 
\def\confName{CVPR}
\def\confYear{2026}

\title{Elucidating the Design Space of Arbitrary-Noise-Based Diffusion Models}

\author{
	Xingyu Qiu$^{1}$, 
	Mengying Yang$^{1}$, 
	Xinghua Ma$^{1}$,
	Dong Liang$^{1}$, Fanding Li$^{1}$,\\
	\textbf{Gongning Luo$^{1*}$},	Wei Wang$^{2}$,	Kuanquan Wang$^{1}$, Shuo Li$^{3}$ \\
	$^{1}$Harbin Institute of Technology, Harbin, China \\
	$^{2}$Harbin Institute of Technology, Shenzhen, China \\
	$^{3}$Case Western Reserve University, Cleveland, USA \\
	\texttt{luogongning@hit.edu.cn}
}

\begin{document}
\maketitle
\begin{abstract}
	Although EDM aims to unify the design space of diffusion models, its reliance on fixed Gaussian noise prevents it from explaining emerging flow-based methods that diffuse arbitrary noise. Moreover, our study reveals that EDM's forcible injection of Gaussian noise has adverse effects on image restoration task, as it corrupts the degraded images, overextends the restoration distance, and increases the task's complexity.
	To interpret diverse methods for handling distinct noise patterns within a unified theoretical framework and to minimize the restoration distance, we propose \textbf{EDA}, which \textbf{E}lucidates the \textbf{D}esign space of \textbf{A}rbitrary-noise diffusion models. Theoretically, EDA expands noise pattern flexibility while preserving EDM's modularity, with rigorous proof that increased noise complexity introduces no additional computational overhead during restoration.
	EDA is validated on three representative medical image denoising and natural image restoration tasks: MRI bias field correction (global smooth noise), CT metal artifact removal (global sharp noise) and natural image shadow removal (local boundary-aware noise). With only 5 sampling steps, competitive results against specialized methods across medical and natural tasks demonstrate EDA's strong generalization capability for image restoration. 
	Code is available at: https://github.com/PerceptionComputingLab/EDA.
\end{abstract}    
\section{Introduction}
\begin{figure}[h!]
	\centering
	\includegraphics[width=1\linewidth]{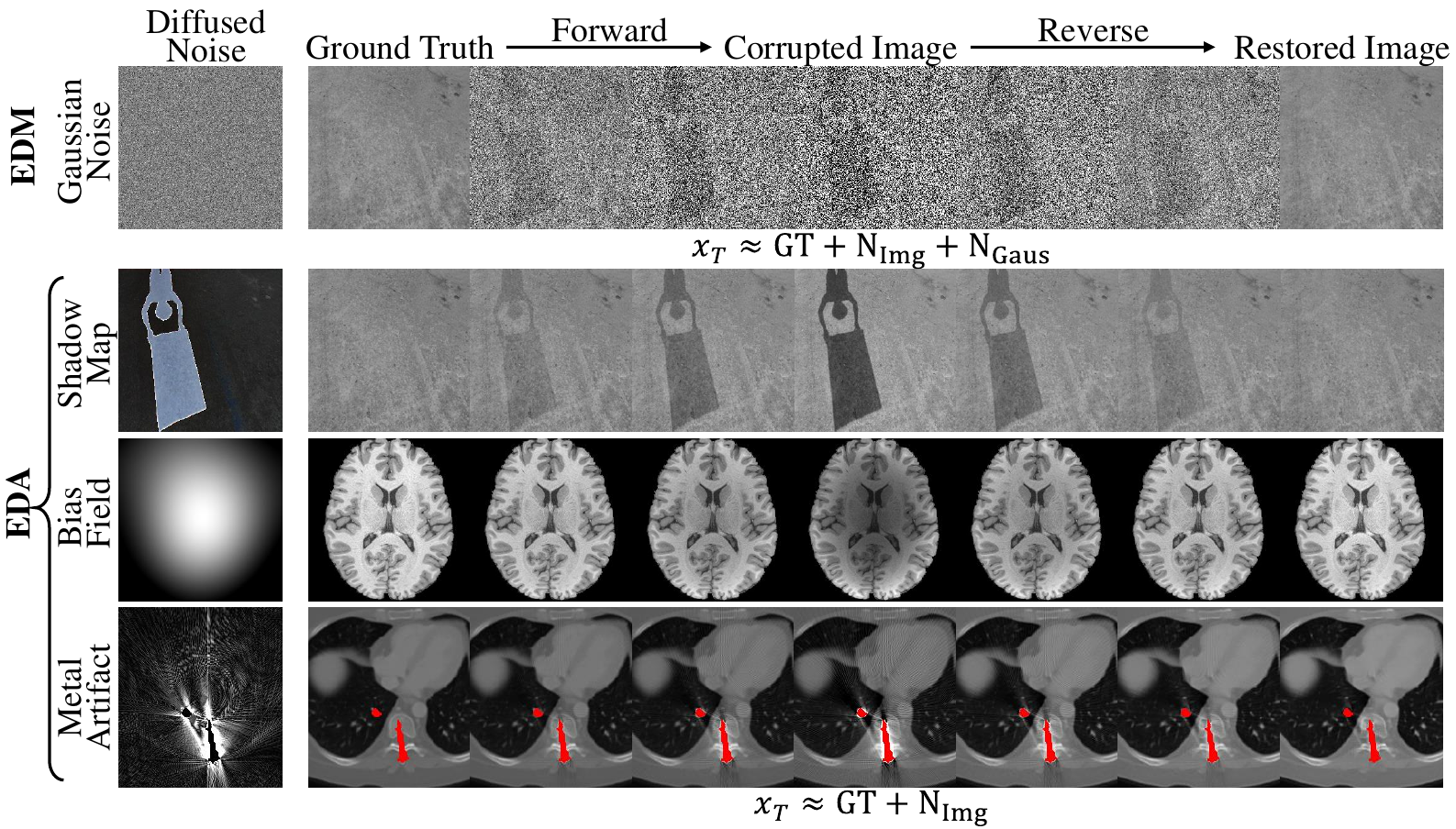} 
	\caption{Our EDA supports arbitrary noise patterns diffusion ($\text{N}_\text{Img}$) and enables the reverse process to be initiated directly from the known degraded image. This avoids the extra Gaussian noise corruption ($\text{N}_\text{Gaus}$; top row) of EDM-based methods, and shortens the image restoration distance, while reducing task complexity and achieving high-quality results with fewer sampling steps.}
	\label{Outputs}
\end{figure}
\begin{figure}[htbp]
	\centering
	\begin{subfigure}{1\columnwidth}
		\includegraphics[width=\linewidth]{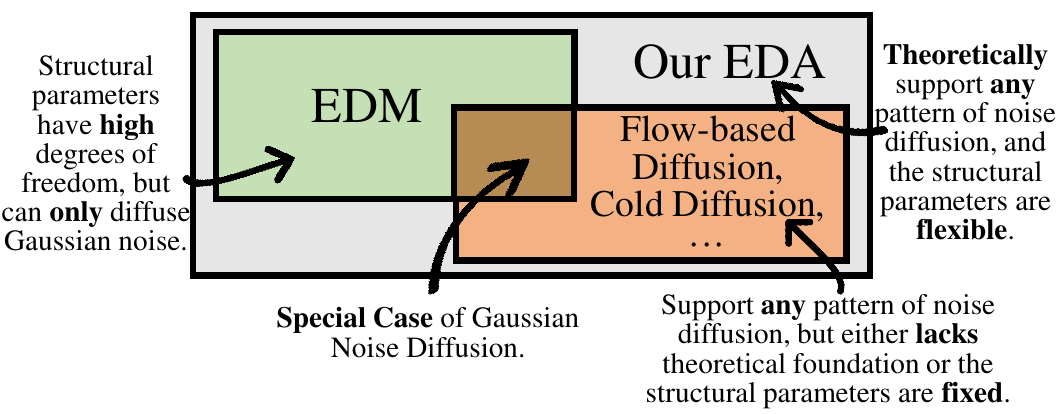}
		\caption{Scope of our EDA.}
		\label{fig:sub1}
	\end{subfigure}
	\hfill 
	\begin{subfigure}{.7\columnwidth}
		\includegraphics[width=\linewidth]{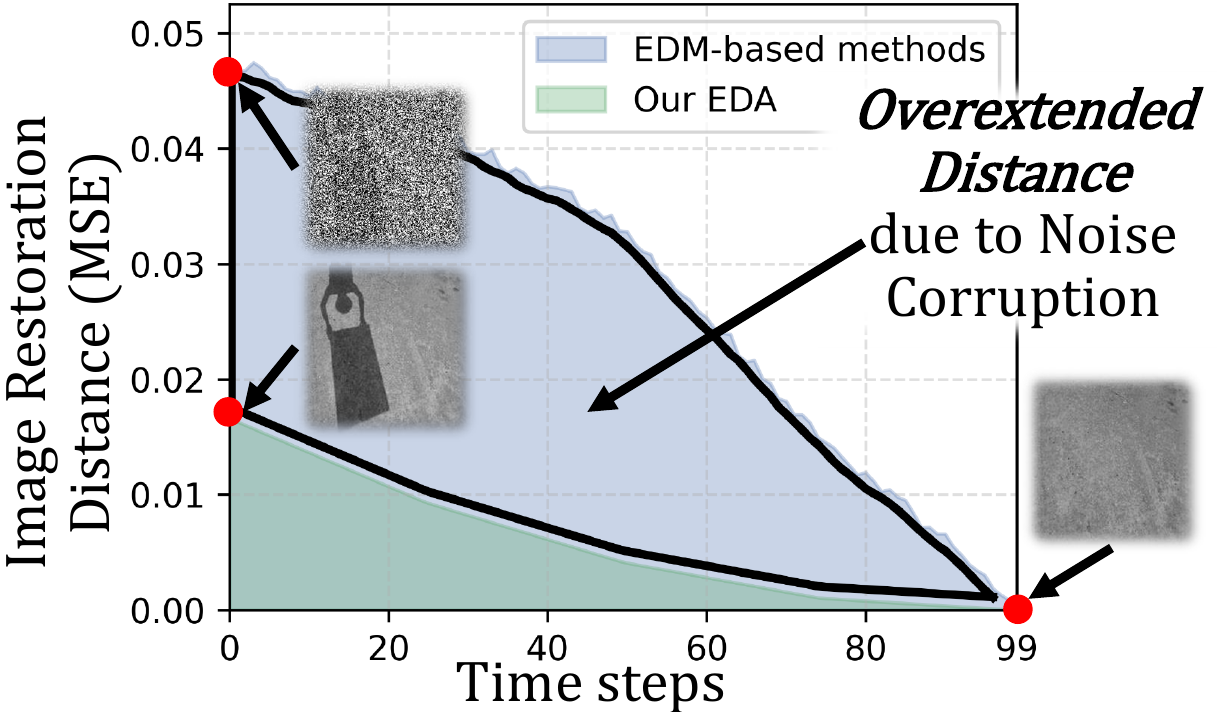}
		\caption{Comparison of restoration distance for diffusion.}
		\label{fig:sub2}
	\end{subfigure}
	\caption{(a) EDA extends EDM by enabling any diffused noise patterns while retaining the flexibility of structural parameters, such as noise schedules and training objectives. (b) EDA enables arbitrary noise diffusion avoiding extra Gaussian noise, and reducing the image restoration distance during the reverse process.}
	\label{fig:overall_fig}
\end{figure}

Currently, there is no unified design space for diffusion models that accommodates both flexible noise patterns and high-performance
SDE-based diffusion, which hinders the theoretical development of more generalized diffusion models.
EDM~\cite{karras2022elucidating} unifies most diffusion models~\cite{DDIM,DDPM,song2019generative,song2020score},
but its restriction to diffusing only Gaussian noise (as shown in Fig.~\ref{fig:sub1}) prevents it from encompassing
more flexible diffusion methods~\cite{bansal2022cold,geng2025meanflow}.
Alternatively, Flow Matching~\cite{lipman2022flow} offers a ODE diffusion framework, breaking
the constraints of Gaussian noise diffusion. This allows more flexible diffusion processes and enables more efficient
sampling. Nevertheless, SDE-based approaches have superior diversity and quality of results as shown in numerous studies \cite{DDIM,song2020score,karras2022elucidating} compared to ODE diffusion.
Consequently, establishing a unified SDE-based design space that enhances the flexibility of noise patterns is important,
which would empower researchers to \emph{articulate the theoretical distinctions} between disparate diffusion methods
from a consistent viewpoint, thereby laying a robust foundation for future diffusion model development.
Addressing this open challenge, the EDA diffusion framework proposed in this paper offers a viable solution.
%
%

High-flexibility diffused noise patterns provide greater advantages for image restoration compared to generation.
When restricted to Gaussian noise, vanilla diffusion-based restoration methods \cite{RDDM,IRSDE,luo2023refusion,kawar2022denoising} initiate the reverse process from noise-corrupted degraded images or pure noise maps, progressively restoring them toward the target domain. However, this forced noise injection introduces two key limitations: 
(1) degraded images lose task-specific information due to added noise corruption, (2) the imposed noise artificially increases the image restoration distance and restoration complexity as shown in Fig. \ref{fig:sub2}.
This distance can be reduced by customizing the diffused noise patterns and initiating the reverse process directly from the known degraded image in the task, thereby \emph{simplifying the restoration}. 
Consequently, the restrictive reliance of the EDM framework on purely Gaussian diffusion noise fundamentally undermines its efficiency and practical utility for high-fidelity image restoration tasks.

This paper \textbf{E}lucidates the \textbf{D}esign space of diffusion models with \textbf{A}rbitrary noise patterns, termed EDA.
EDA interprets diffusion methods handling diverse noise patterns within a unified theoretical framework as shown in Fig. \ref{Outputs}, simplifying and unifying diffusion-based restoration processes originally developed for Gaussian noise constraints.
Specifically, EDA characterizes the diffusion process through multivariate Gaussian distributions, and our study theoretically demonstrates its capacity to diffuse arbitrary noise patterns. The framework models diffused noise via multivariate Gaussian distributions and derives a corresponding stochastic differential equation (SDE) driven by multiple independent Wiener processes. Deterministic sampling rules are derived by solving the probability flow ordinary differential equation (PFODE).
Our study indicates that EDA incurs no additional computational cost when generalizing from simple Gaussian noise to complex patterns.
EDA is evaluated on three restoration tasks, including two critical medical image denoising applications: MRI bias correction (global smooth noise) and CT metal artifact reduction (global sharp noise), along with natural image shadow removal (local boundary-aware noise), which is commonly used. Superior experimental results on both medical and natural images demonstrate EDA's strong generalization ability for image restoration.
With fewer than $5$ sampling steps, EDA performs competitively with $100$-step EDM-based methods, outperforms task-specific methods, and establishes state-of-the-art results in bias field correction and shadow removal.

\section{Related Work}
\subsection{EDM can Only Diffuse Gaussian Noise}
EDM \cite{karras2022elucidating} elucidates the unified design space of diffusion-based generative models, and it has significant flexibility of structural parameters, such as noise schedules and training objectives. But EDM is limited by a single diffused noise pattern, pure Gaussian noise. The forward process is defined by a stochastic differential equation (SDE):
\begin{equation}
	\label{1}
	\mathrm{d}\bm{x} = f(t)\bm{x}\,\mathrm{d}t + g(t)\,\mathrm{d}\omega_t
\end{equation}
where $x \in \mathbb{R}^d$ represents the data sample at time $t$, $f(t): \mathbb{R} \to \mathbb{R}$ and $g(t): \mathbb{R} \to \mathbb{R}$ are drift and diffusion coefficients, and $\omega_t$ denotes the standard Wiener process. 
The signal scaling function $s(t)$ and noise level function $\sigma(t)$ represent the noise schedules, which are derived as $s(t) = e^{\int_0^t f(r)\,\mathrm{d}r}$ and $\sigma(t) = \sqrt{\int_0^t \frac{g^2(r)}{s^2(r)}\,\mathrm{d}r}$. This leads to a perturbed distribution at time $t$:
\begin{equation}
	\label{3}
	p_t(\bm{x}_t \mid \bm{x}_0) = \mathcal{N}\left(\bm{x}_t;\, s(t)\bm{x}_0\,, s^2(t)\sigma^2(t)\mathbf{I}\right)
\end{equation}
where $x_0 \sim p_{data}$ is the clean data sample, $\mathcal{N}(\cdot)$ denotes the Gaussian distribution, and $\mathbf{I}$ is the identity matrix. The deterministic probability flow ordinary differential equation (PFODE) is expressed as:
\begin{equation}
	\label{4}
	\mathrm{d}\bm{x} = \left[\frac{s'(t)}{s(t)}\bm{x} - s^2(t)\sigma'(t)\sigma(t)\nabla_{\bm{x}}\log p\left(\frac{\bm{x}}{s(t)}; \sigma(t)\right)\right]\mathrm{d}t
\end{equation}
where $s'(t)\triangleq \mathrm{d}\, s(t) / \mathrm{d}\, t$, $\sigma'(t)\triangleq \mathrm{d}\, \sigma(t) / \mathrm{d}\, t$, and the score function $\nabla_{\bm{x}}\log p(\cdot)$ is parameterized by a denoiser $D(\bm{x};\sigma)$:
\begin{equation}
	\label{5}
	\nabla_{\bm{x}}\log p(\bm{x};\sigma) = \frac{D\left(\dfrac{\bm{x}}{s(t)};\sigma\right) - \dfrac{\bm{x}}{s(t)}}{s(t)\sigma^2(t)}
\end{equation}
The EDM trains the denoiser $D(\bm{x};\sigma)$ by:
\begin{equation}
	\label{666}
	\mathcal{L} = \mathbb{E}_{\bm{x}_0 \sim P_{data}}\mathbb{E}_{\bm{x} \sim P(\bm{x}_t | \bm{y})} \lVert D_\theta(\bm{x}; \sigma) - \bm{x}_0 \rVert^2
\end{equation}
The $D(\bm{x};\sigma)$ is defined as:
\begin{equation}
	\label{66}
	D_\theta(\bm{x};\sigma)=c_{skip}(\sigma) \bm{x}+c_{out}(\sigma) F_\theta(c_{in}(\sigma)\bm{x};c_{noise}(\sigma))
\end{equation}
where $F_\theta$ is the trainable neural network, $c_{skip}(\sigma)$, $c_{out}(\sigma)$, $c_{in}(\sigma)$ and $c_{noise}(\sigma)$ are hyperparameters. The $c_{skip}(\sigma)$ and $c_{out}(\sigma)$ depend on the training objective of $F_\theta$ and noise addition strategy $f(\cdot), g(\cdot)$. The $c_{in}(\sigma)$ scales the input data to a normalized interval, and $c_{noise}(\sigma)$ encodes time information.

Substituting the score function Eq. \ref{5} into PFODE Eq. \ref{4} yields the final deterministic sampling rule:
\begin{equation}
	\label{77}
	\frac{\mathrm{d}\bm{x}}{\mathrm{d}t} = \left( \frac{s'(t)}{s(t)} + \frac{\sigma'(t)}{\sigma(t)} \right) \bm{x} - \frac{\sigma'(t)s(t)}{\sigma(t)} D_\theta\left(\dfrac{\bm{x}}{s(t)}; \sigma\right)
\end{equation}

Above all, the noise schedules $s(\cdot), \sigma(\cdot)$ in Eq. \ref{3} and the training objectives of $F_\theta$ in Eq. \ref{66} are arbitrary. 
This high degree of flexibility allows the framework to unify nearly all diffusion-based generation methods. 
For example, DDPM \cite{DDPM} and DDIM \cite{DDIM} are spacial cases of EDM when $s(t)=\sqrt{\bar{\alpha_t}}$, $\sigma(t)=\sqrt{1-\bar{\alpha_t}}/\sqrt{\bar{\alpha_t}}$, and the training objectives of $F_\theta(x_t,t)$ is the noise map $\epsilon$ in $x_t$.

While EDM offers theoretical unification, its single diffused noise pattern introduces critical limitations. (1) The closed-form perturbation kernel in Eq. \ref{3} limits diffusion to pure Gaussian noise characterized by pixel-wise independence and homogeneity, precluding support for arbitrary noise patterns. (2) In restoration tasks where the initial state $x_T$ is the degraded image $P_{LQ}$, however EDM-based methods \cite{IRSDE,kawar2022denoising,luo2023refusion,RDDM,liu2023i2sb,delbracioinversion,li2023bbdm,zhengdiffusion,dou2025image,shi2023diffusion} introduces an artificial restoration gap as its initial state must include Gaussian noise $P_{LQ}+ N_{Gaus}$. Our EDA addresses this by increasing the flexibility of diffused noise patterns while retaining the flexibility of structural parameters.

\subsection{Diffusion-based Methods for Any Noise Diffusion}
Recent advances in diffusion models have extended diffused noise from pure Gaussian noise to arbitrary noise patterns, but this comes at the cost of theoretical simplification, forcing the generation process to degrade from SDE to deterministic ODEs, or even lacking a clear theoretical grounding, which fundamentally prevents the EDM design space from encompassing these new approaches.
Flow Matching \cite{lipman2022flow} enables transformations between arbitrary distributions by constructing continuous probability paths, directly modeling deterministic transport from an initial noise distribution $p_1(x)$ to a target data distribution $p_0(x)$.
Its key innovation lies in designing vector fields via continuous normalizing flows (CNFs), allowing samples $x_0\sim p_0$ to evolve along these paths toward $x_1$.
The core dynamics are defined via ODE:
\begin{equation}
\frac{d x_t}{d t} = v_t(x_t)
\end{equation}
where $x_t$ denotes the sample state at time $t\in[0,1]$, and $v_t(x_t)$ is the time-dependent velocity field learned by the model to drive this evolution.
MeanFlow \cite{geng2025meanflow} extends Flow Matching by characterizing the flow field using the notion of average velocity, in contrast to the instantaneous velocity modeled by its predecessor. 
This theoretical refinement is crucial for efficiency, enabling MeanFlow to achieve state-of-the-art performance in one-step generative modeling, making it a critical benchmark for highly efficient diffusion and flow methods.
Cold Diffusion \cite{bansal2022cold} sacrifices the theoretical grounding in stochastic processes to gain enhanced flexibility and generality across tasks. It does so by replacing Gaussian noise with deterministic degradation operators $D(\cdot)$ and training networks to invert them via restoration functions $R(\cdot)$.

These works collectively propose diverse diffusion processes, demonstrating that the strength of diffusion models lies in their progressive transformation framework, not specific noise types.
Despite the high flexibility of ODE diffusions, they may exhibit reduced robustness due to the lack of stochasticity \cite{DDIM,song2020score,karras2022elucidating}, so the development of a unified framework that retains the SDE design space is essential for leveraging its advantages in generating high-fidelity and diverse samples.
\emph{Following previous work, this paper denotes these diffusion mediums as \textbf{generalized noise}}.
Our EDA elucidates a comprehensive design space that encompasses both SDE-based diffusion models and diverse, generalized diffusion processes, thereby avoiding the restrictive Gaussian noise assumption inherent in EDM.

\section{Methods}
This paper proposes a novel design space EDA that generalizes the vanilla diffusion design space (EDM \cite{karras2022elucidating}) to enable arbitrary noise patterns while maintaining the flexibility of structural parameters.
Specifically, Sec. \ref{3.1} and \ref{3.2} introduce the theoretical framework EDA based on the multivariate Gaussian distribution, which increases the flexibility of diffuse noise patterns. Subsequently, Prop. \ref{prop1} proves that EDA has the capability to achieve arbitrary noise diffusion.
\subsection{Generalized Forward Process}
\label{3.1}
Vanilla diffusion models rely on pixel-wise independent Gaussian noise, characterized by simplicity and ease of modeling, yet it inherently limits the diversity of diffusion patterns. In contrast, our EDA introduces novel diffused noise patterns with high degrees of freedom and diverse diffusion processes through an unconstrained covariance parameterized by arbitrary basis functions and stochasticity weights.

Let $x_0 \sim P_{data}$ be the data distribution. The diffused noise $N$ in EDA is defined as:
\begin{equation}
	\label{6}
	N=\sum_{m=1}^M \dfrac{\eta+\epsilon_m}{\eta+1}h_{m,\bm{x}_0}
\end{equation}
where $H_{\bm{x}_0} = [h_{1,\bm{x}_0},...,h_{M,\bm{x}_0}]$ are basis set that adjust noise patterns, each $h_{m,\bm{x}_0}$ is dimensionally aligned with $x_0$, the $\epsilon_1,...,\epsilon_M\sim \mathcal{N}(0,1)$ are independent Gaussian variables, and $\eta \geq 0$ is a mediator for adjusting the stochasticity of diffused noise. 
The basis functions $H_{x_0}$ can be predefined arbitrarily and do not require strict orthogonality (see Appendix B), enabling flexible noise modeling. The parameter $\eta$ controls the noise stochasticity: when $\eta = 0$, the noise is maximally stochastic; as $\eta \rightarrow + \infty$, the noise becomes deterministic.

The generalized forward process is formulated as:
\begin{equation}
	\label{7}
	\bm{x}_t = s(t)\bm{x}_0 + s(t)\sigma(t) \sum_{m=1}^M \frac{\epsilon_m + \eta}{\eta + 1} h_{m,\bm{x}_0}
\end{equation}
where $s(t)$ scales the signal and $\sigma(t)$ schedules the noise intensity. The definitions of $s(t)$ and $\sigma(t)$ are the same as in Eq. \ref{3}, except that the diffused noise $N$ is replaced from Gaussian noise to Eq. \ref{6}. The corresponding distributions are:
\begin{equation}
    \label{8}
    \begin{aligned}
    P(\bm{x}_t \mid \bm{x}_0)
      &= \mathcal N\bigl(\bm{x}_t;\, \mu_t,\, \sigma_t\bigr),\\
    \mu_t
      &= s(t)\,\bm{x}_0
       + \frac{\eta\, s(t)\, \sigma(t)}{\eta + 1}
         \sum_{m=1}^{M} h_{m,\bm{x}_0},\\
    \sigma_t
      &= \frac{s^2(t)\, \sigma^2(t)}{(\eta + 1)^2}\, \bm{\Sigma}_{\bm{x}_0}
    \end{aligned}
\end{equation}
with covariance $\bm{\Sigma}_{\bm{x}_0}=H_{\bm{x}_0}H_{\bm{x}_0}^\top$. Unlike EDM’s diagonal covariance in Eq. \ref{3}, EDA’s $\bm{\Sigma}_{\bm{x}_0}$ captures structured noise through the predefined basis set $H_{\bm{x}_0}$, enriching the noise patterns and perturbation space.

Corresponding to the generalized forward process Eq. \ref{8} of EDA, our forward process is controlled by an SDE driven by multiple independent Wiener processes:
\begin{equation}
	\label{9}
	\mathrm{d}\bm{x} = \left[ f(t)\bm{x} + \phi_{\bm{x}_0}(t) \right] \mathrm{d}t + g(t) \sum_{m=1}^M h_{m,\bm{x}_0} \mathrm{d}\omega_t^{(m)}
\end{equation}
where $f(\cdot),g(\cdot): \mathbb{R} \to \mathbb{R}$ and $\phi(\cdot): \mathbb{R} \to \mathbb{R}^d$, $\mathrm{d}\omega_t^{(1)},...,\mathrm{d}\omega_t^{(M)}$ are independent Wiener increments. The terms $f(t)$, $g(t)$ and $\phi_{\bm{x}_0}(t)$ directly map to the scale $s(t)$, noise schedule $\sigma(t)$ and basis set $H_{\bm{x}_0}$ (see Appendix A), which ensure that the SDE of Eq. \ref{9} is consistent with the forward process Eq. \ref{8}:
\begin{equation}
	\label{10}
	\begin{split}
		f(t) = \frac{s'(t)}{s(t)},& \quad g(t) =\frac{s(t)}{\eta + 1} \sqrt{\frac{\mathrm{d}\sigma^2(t)}{\mathrm{d}t}}\\
		\phi_{\bm{x}_0}(t) =& \frac{\eta s(t) \sigma'(t)}{\eta + 1} \sum_{m=1}^M h_{m,\bm{x}_0}\\
	\end{split}
\end{equation}

Our EDA generalizes the noise diffusion beyond pure Gaussian noise, and our SDE framework supports various noise diffusion, broadening applicability to complex corruption processes.

\subsection{Deterministic Sampling via PFODE}
\label{3.2}
This section introduces the Probabilistic Flow Ordinary Differential Equation \cite{song2020score} (PFODE) for the EDA, which shares the same solution space as the SDE Eq. \ref{9} with diverse diffusion processes. Then, the deterministic sampling formulation of EDA is derived based on PFODE. The defining feature of our PFODE is that evolving a sample $x_a\sim P(x_a|x_0)$ from time $a$ to $b$ (time forward or backward) produces a sample $x_b\sim p(x_b|x_0)$. The PFODE for EDA is derived as (see Appendix B):
\begin{equation}
	\label{11}
	\frac{\mathrm{d}\bm{x}}{\mathrm{d}t} = f(t)\bm{x} + \phi_{\bm{x}_0}(t) - \frac{1}{2} g^2(t) \bm{\Sigma}_{\bm{x}_0} \nabla_{\bm{x}} \log P(\bm{x} | \bm{x}_0)
\end{equation}
where the specific forms of $f(t)$, $g(t)$ and $\phi_{\bm{x}_0}(t)$ are given in Eq. \ref{10}. The $\nabla_{\bm{x}} \log P(\bm{x} | \bm{x}_0)$ is defined as score function \cite{song2020score}, which is derived from the probability density function of the distribution Eq. \ref{8} (see Appendix B):
\begin{equation}
    \label{12}
    \begin{aligned}
    &\nabla_{\bm{x}} \log P(\bm{x} \mid \bm{x}_0)
    = \frac{(\eta + 1)^2}{s^2(t)\sigma^2(t)}\, \bm{\Sigma}_{\bm{x}_0}^{-1} \\
    &\quad \cdot \Bigl(
         s(t)\,\bm{x}_0
         + \frac{\eta\, s(t)\, \sigma(t)}{\eta + 1}
           \sum_{m=1}^{M} h_{m,\bm{x}_0}
         - \bm{x}
       \Bigr)
    \end{aligned}
\end{equation}
Computing the score function $\nabla_{\bm{x}} \log P(\bm{x} | \bm{x}_0)$ with a denoiser $D_\theta$ and substituting $f(t)$, $g(t)$ and 
$\phi_{\bm{x}_0}(t)$ with Eq. \ref{10} into the PFODE Eq. \ref{11} yields the deterministic update rule (see Appendix C):
\begin{equation}
	\label{13}
	\frac{\mathrm{d}\bm{x}}{\mathrm{d}t} = \left( \frac{s'(t)}{s(t)} + \frac{\sigma'(t)}{\sigma(t)} \right) \bm{x} - \frac{\sigma'(t)s(t)}{\sigma(t)} D_\theta(\bm{x}; \sigma)
\end{equation}
The denoiser $D_\theta(\bm{x}; \sigma)$ is trained via:
\begin{equation}
	\label{14}
	\mathcal{L} = \mathbb{E}_{\bm{x}_0 \sim P_{data}}\mathbb{E}_{\bm{x} \sim P(\bm{x}_t | \bm{y})} \lVert D_\theta(\bm{x}; \sigma) - \bm{x}_0 \rVert^2
\end{equation}
where the definition of $D_\theta(\bm{x}; \sigma)$ is same as Eq. \ref{66} as:
\begin{equation}
	D_\theta(\bm{x};\sigma)=c_{skip}(\sigma) \bm{x}+c_{out}(\sigma) F_\theta(c_{in}(\sigma)\bm{x};c_{noise}(\sigma))
\end{equation}
where the $F_\theta(x_t,t)$ has flexible training objectives such as clean image, diffused noise \cite{DDPM,DDIM}, velocity field \cite{lipman2022flow}, score function \cite{song2020score}. The final training process for the EDA is provided in Algorithm \ref{train}. This paper uses the same training objectives as DDPM \cite{DDPM} and DDIM \cite{DDIM}, whose networks predict diffused noise, $F_\theta(\cdot)\approx N$.
Compared to Heun’s $2^{\text{nd}}$ method or higher-order solvers, Euler’s $1^{\text{st}}$ method is less computationally intensive and remains widely adopted due to its simplicity \cite{DDPM,DDIM,song2020score,lipman2022flow}. Consequently, Algorithm \ref{euler_sampler} provides a deterministic sampling procedure for EDM based on Euler’s $1^{\text{st}}$ method.

\begin{algorithm}
	\caption{The EDA training process with arbitrary diffused noise pattern $N$.}
	\label{train}
	\begin{algorithmic}[1]
		\Require{$P_{data}$, $H$, $\eta$, $\sigma(t)$, $s(t)$, time training strategies $P_{time}$, training step $N$,  }
		\Statex
		\State Initialize the neural network $F_\theta(\cdot)$
		\For{training step $s \in \{1, \ldots, N\}$}
		\State Sample data $\bm{x}_0 \sim P_{data}$
		\State Sample time $t \sim P_{time}$
		\State Get basis set $H_{\bm{x}_0} = [h_{1,\bm{x}_0},...,h_{M,\bm{x}_0}]$
		\State Sample diffusion data: $\bm{x}_t\sim$ 
		\vspace{-20pt}\Statex \[
				\mathcal{N}\!\left(
				s(t)\bm{x}_0
				+ \frac{\eta\, s(t)\, \sigma(t)}{\eta + 1}
				\sum_{m=1}^{M} h_{m,\bm{x}_0},\;
				\frac{s^2(t)\sigma^2(t)}{(\eta + 1)^2}
				\bm{\Sigma}_{\bm{x}_0}
			\right)
		\]
		\State Compute loss function:
		$\mathcal{L} = \lVert D_\theta(\bm{x}; \sigma) - \bm{x_0} \rVert^2$
		\State Update network parameters with $\nabla_{\theta}\mathcal{L}$
		\EndFor
		\State \Return $D_{\theta}(\cdot)$ \Comment{Return trained denoiser}
	\end{algorithmic}
\end{algorithm}

\begin{Proposition}
	\label{prop1}
	EDA Supports Diffusion and Removal of Arbitrary Noise.
\end{Proposition}

\emph{Proof.}
The EDA achieves adaptability to arbitrary noise patterns through an unconstrained  basis set $H_{\bm{x}_0} = [h_{1,\bm{x}_0},...,h_{M,\bm{x}_0}]$, which controls the covariance of diffusion process. Below, three representative cases are formally presented:
\begin{enumerate}
	\item \emph{Unified Basis Set (Optimal Case).} This case represents the optimal configuration for EDA (see Appendix D) when noise can be decomposed into a fixed basis set $H$, independent of the data sample $x_0$, such that $H=H_j$ for $\forall j \sim P_{data}$ and the covariance $\Sigma=HH^\top$ remains constant across samples.  For example, in MRI bias field correction (Fig. \ref{Outputs}), the smooth bias field can be represented by a series of low-frequency basis functions $H=[h_1,...,h_M]$, where each $h_m$ is a smooth template independent of $x_0$ (Appendix E.2). However, the assumption of a fixed basis set is restrictive, limiting practical applicability.
	
	\item \emph{Sample-Dependent Basis Functions (General Case).} This case applies universally. When noise patterns cannot be decomposed into a fixed basis set, $H_{\bm{x}_0}$ becomes sample-dependent. Take a simple example, mapping an image $\bm{A}$ to its degraded counterpart $\bm{B}$, where the basis set reduces to $H_{\bm{A}}=[\bm{B}-\bm{A}]$. Unlike Case 1, Case 2 uses the approximately optimal output of the denoiser in the deterministic sampling process (see Appendix C).	Applications include metal artifact reduction (MAR) in CT and shadow removal in natural images (Appendices E.3, E.4).
	
	\item \emph{Non-Gaussian Noise via Discrete Sampling.} Though this case is a special instance of Case 2, it addresses potential confusion arising from non-Gaussian noise distributions $P_{noise}$ (like Poisson noise). By discretizing experimental scenarios, assuming a discrete sample $N_1 \sim P_{noise}$, and our diffused noise is constructed as $N_2=\frac{\epsilon + \eta}{\eta + 1} \bm{h}$ where $\bm{h}\sim P_{noise}$ and $\epsilon \sim \mathcal{N}(0,1)$. By adjusting an appropriate $\eta$, $N_1$ and $N_2$ share identical distributions in discrete space, enabling EDA to support random noise other than Gaussian in practice.
\end{enumerate}

\begin{algorithm}
	\caption{Deterministic sampling using Euler's method.}
	\label{euler_sampler}
	\begin{algorithmic}[1]
		\Require{$D_{\theta}(x;\sigma)$, $\sigma(t)$, $s(t)$, time steps $t_{i\in\{0,\ldots,N\}}$}
		\Statex
		\State Get initial sample $x_{T} \sim P_{LQ} \approx P(\bm{x}_T | \bm{x}_0)$
		\For{$i \in \{T, \ldots, 1\}$}
		\State Evaluate $\mathrm{d}x/\mathrm{d}t$ at $t_i$:
		\[
		d_{i} \leftarrow \left( \frac{s'(t)}{s(t)} + \frac{\sigma'(t)}{\sigma(t)} \right) \bm{x}_t - \frac{\sigma'(t)s(t)}{\sigma(t)} D_\theta(\bm{x}; \sigma)
		\]
		\State Update with Euler step:
		\[
		x_{i-1} \gets x_{i} + (t_{i-1} - t_{i})d_{i}
		\]
		\EndFor
		\State \Return $x_{0}$ \Comment{Return noise-free sample} 
	\end{algorithmic}
\end{algorithm}

\begin{Proposition}
	High Flexibility of Diffused Noise Pattern Does Not Increase Sampling Complexity.
\end{Proposition}
\emph{Proof.}
EDA aligns with EDM in deterministic sampling and denoiser training objective. While the diffusion process in EDA incorporates an arbitrary covariance matrix $\Sigma_{\bm{x}_0}$, contrasting with EDM’s diagonal covariance $\sigma^2(t)\mathbf{I}$, this introduces additional terms in the $P(\bm{x}_t | \bm{x}_0)$ (Eq. \ref{3} vs. \ref{8}), the SDE (Eq. \ref{1} vs. \ref{9}), the PFODE (Eq. \ref{4} vs. \ref{11}), and the score functions (Eq. \ref{5} vs. \ref{12}). Remarkably, these additional terms can be analytically simplified and eliminated (Appendix C), resulting in our deterministic sampling formulation Eq. \ref{13}) that aligns with EDM (Eq. \ref{77}). Furthermore, the denoiser is trained using the same objective, as shown in Eq. \ref{666} and \ref{14}.

\begin{Proposition}
	EDM is a Special Case of Our EDA.
\end{Proposition}
\emph{Proof.}
EDM \cite{karras2022elucidating} is a strict subset of EDA under specific parameter configurations as shown in Fig. \ref{fig:sub1}. When the parameters are set to: (1) the mediator $\eta=0$; (2) the basis set $H = [E_{1, 1},..., E_{H, W}]$, where each $E_{i,j}$ is a matrix with a value of $1$ at position $(i,j)$ and $0$ elsewhere. The diffused noise $N$ in EDA Eq. \ref{6} reduces to pixel-wise independent Gaussian noise:
\begin{equation}
	\bm{N} = \sum_{i,j} \epsilon_{i,j} E_{i,j}, \quad \epsilon_{i,j} \sim \mathcal{N}(0,1)
\end{equation}
This is the same noise pattern as in EDM. Consequently, the diffusion process simplifies to:
\begin{equation}
	\bm{x}_t = s(t)\bm{x}_0 + s(t)\sigma(t) \bm{N}, \quad \bm{N} \sim \mathcal{N}(0, \mathbf{I})
\end{equation}
which is identical to EDM’s additive Gaussian corruption Eq. \ref{3}. The forward process and deterministic sampling further align with EDM, respectively.

\section{Experiments}
\begin{figure*}[!t]
	\centering
	\includegraphics[width=.92\textwidth]{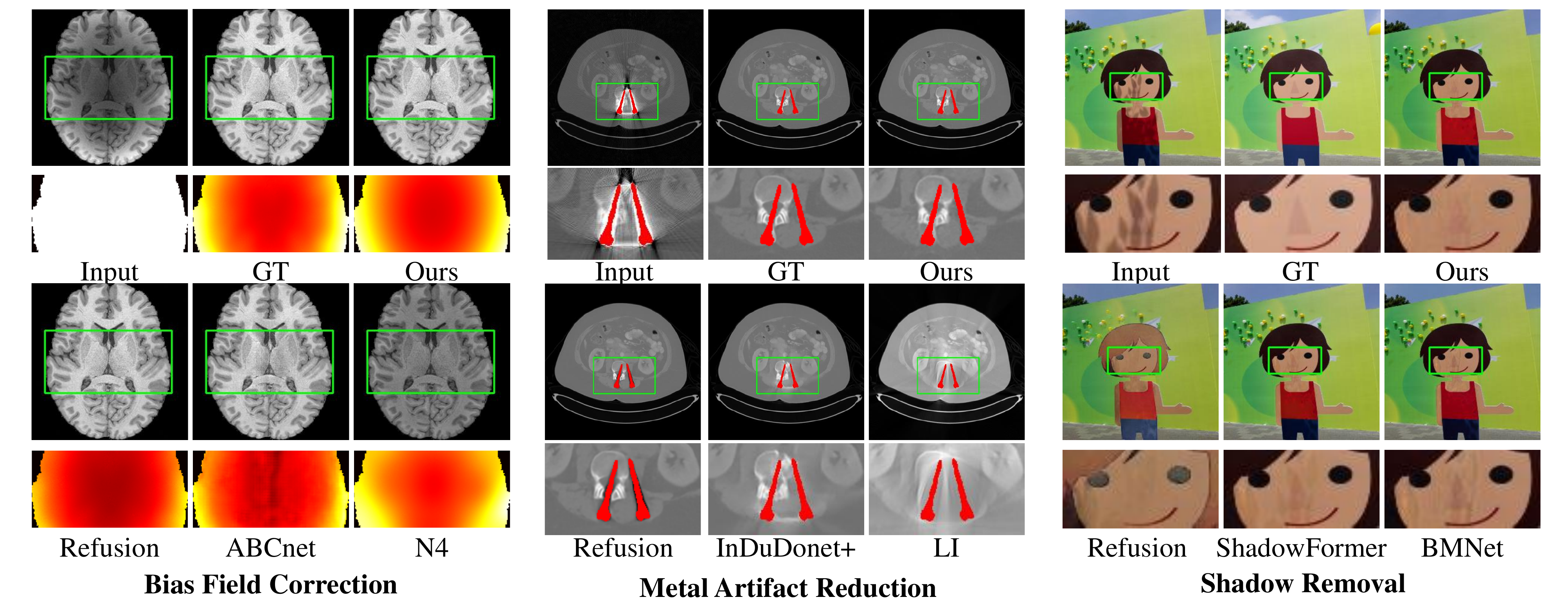} 
	\caption{Our EDA produces the closest results to the ground truth and has competitive image restoration performance. The enlarged view of the green box in the bias field correction is the bias field.}
	\label{results1}
\end{figure*}
\begin{figure}[!t] 
	\centering
	\includegraphics[width=0.5\textwidth]{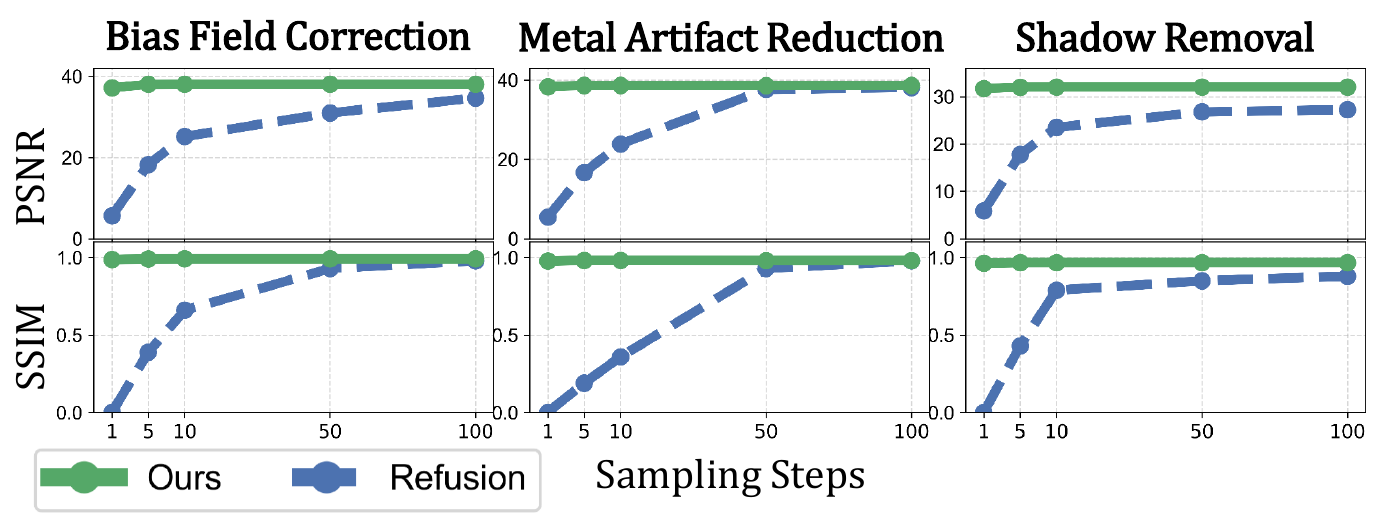}
	\caption{Our EDA samples in less than $5$ steps achieve or even surpass the Refusion sampling in $100$ steps.}
	\label{steps}
\end{figure}
Our EDA has been evaluated on three distinct and representative image restoration tasks.
With fewer than $5$ sampling steps, EDA outperforms most task-specific methods and achieves state-of-the-art performance in both bias field correction and shadow removal.

\subsection{Tasks and Datasets}
\label{task}
Three distinct restoration tasks were chosen for evaluation, with representative noises as shown in Fig. \ref{Outputs}: (1) global smooth noise, (2) global sharp noise, (3) local noise with boundaries. 

\emph{Task 1: Bias field correction (BFC)} is a crucial step in MRI processing. 
The removed bias field is globally smooth and serves to homogenize MRI intensities, enhance image quality, and improve diagnostic accuracy. 
This paper uses the clinical dataset from the Human Connectome Project (HCP) \cite{HCP}.
After preprocessing, there are $2206$ slices for training and validation, and $1000$ slices for testing.\\
\emph{Task 2: Metal artifact reduction (MAR)} is an inevitable task in CT scans of patients with metal implants. The metals cause projection data loss and global sharp streak artifacts in CT, affecting the clinical analysis \cite{park2018ct}. 
Following \cite{liao2019adn,lin2019dudonet,lyu2020encoding,yu2020deep,zhang2018convolutional}, our study simulate metal artifacts from DeepLesion \cite{yan2018deep} using $100$ metal masks \cite{zhang2018convolutional}. The training set is $1000$ images with $90$ masks and the test set is $200$ images with $10$ masks.\\
\emph{Task 3: Shadow removal (SR)} aims to restore the local shadow regions with boundaries, improving image quality and benefiting computer vision tasks \cite{sanin2010improved,zhang2018improving}. 
This paper uses the ISTD dataset \cite{wang2018stacked} with $1330$ training and $540$ testing images.

\begingroup
\setlength{\textfloatsep}{6pt}   
\setlength{\floatsep}{6pt}       
\renewcommand{\topfraction}{0.95}
\renewcommand{\floatpagefraction}{0.9}
\begin{table*}[!t]
	\centering
	\small
	\setlength{\tabcolsep}{3pt} 
	\caption{
		The EDA results are closest to the ground truth and outperform existing methods in MRI quality and fidelity.}
	\begin{tabular}{c|ccccc} 
		\toprule
		\multicolumn{1}{c}{Metrics} & \multicolumn{1}{c}{SSIM ↑} & \multicolumn{1}{c}{PSNR ↑}& \multicolumn{1}{c}{COCO ↑}  & \multicolumn{1}{c}{CV(GM) ↓}& \multicolumn{1}{c}{CV(WM) ↓}\\
		\midrule																		
		Input &	0.90 	±	0.08 	&	21.89 	±	11.41 &	0.92 	±	0.06  	&	25.68 	±	6.50 	&	21.37 	±	6.95\\																						
		
		N4  &	0.95 	±	0.05	& 25.62 	±	6.24	&	0.95	±	0.06 &		13.46 	±	1.72 	&	7.95 	±	1.26 
		\\		
		MICO  &	0.94 	±	0.06 	&	23.04 	±	8.21 	& 0.95	±	0.06	&	13.92 	±	1.78 	&	8.49 	±	1.43 \\			
		LapGM  &	0.88 ±	0.04 	&	24.52 	±	5.99 &0.93 	±	0.04 	&		13.02 	±	1.65 	&	\textbf{7.52 	±	1.38} \\																						
		ABCNet &	0.98 	±	0.03 	&	29.58 	±	9.43 &	0.97 	±	0.02 	&	12.74 	±	1.81 	&7.69 	±	1.33  \\		
		MeanFlow & 0.95 ± 0.06   & 26.31 ± 6.88 & 0.94 ± 0.07 & 15.49 ± 1.88 & 7.64 ± 2.21 \\
		Refusion  & \textbf{0.98 	±	0.01 }	&	\textbf{34.67 	±	7.45} 	&	\textbf{0.98 ± 0.01} &	\textbf{12.49 	±	1.67} 	&	7.72 	±	1.56 \\												
		Ours &\color{red}\textbf{0.99 	±	0.01} 	&	\color{red}\textbf{38.02 	±	6.53 } 	&\color{red}\textbf{	0.99 	±	0.02 }
		&	\color{red}\textbf{12.36 	±	1.25 } 	&	\color{red}\textbf{7.40 	±	1.34} \\																																						

		\bottomrule
	\end{tabular}
	\label{BC}
\end{table*}

\begin{table*}[!t]
	\centering
	\small
	\setlength{\tabcolsep}{1.8pt}
	\caption{Our EDA effectively reduces metal artifacts in CT. Using only the image-domain, EDA outperforms partial dual-domain (sinusoidal and image domain) methods, demonstrating the competitive restored performance (PSNR/SSIM).}
	\begin{tabular}{@{}>{\centering}p{8pt}@{\hspace{4pt}}c|ccccc|c@{}} 
		\toprule
		\multicolumn{2}{@{}c}{Methods} & \multicolumn{2}{c}{Large Metal} & \multicolumn{1}{c}{$\rightarrow$} & \multicolumn{2}{c}{Small Metal} & \multicolumn{1}{c}{Average} \\
		\midrule
		
		\multicolumn{2}{@{}c|}{Input}  & 24.12/0.6761 & 26.13/0.7471 & 27.75/0.7659 & 28.53/0.7964 & 28.78/0.8076 & 27.06/0.7586\\
		\midrule
		\multirow{6}{*}{\footnotesize{\rotatebox[origin=c]{90}{\makecell[c]{Dual-domain}} }}
		&LI  & 27.21/0.8920 & 28.31/0.9185 & 29.86/0.9464 & 30.40/0.9555 & 30.57/0.9608 & 29.27/0.9347\\
		&CNNMAR& 28.92/0.9433 & 29.89/0.9588 & 30.84/0.9706 & 31.11/0.9743 & 31.14/0.9752 & 30.38/0.9644\\		
		&DuDoNet  & 29.87/0.9723 & 30.60/0.9786 & 31.46/0.9839 & 31.85/0.9858 & 31.91/0.9862 & 31.14/0.9814\\
		&DSCMAR & 34.04/0.9343 & 33.10/0.9362 & 33.37/0.9384 & 32.75/0.9393 & 32.77/0.9395 & 33.21/0.9375\\
		&InDuDoNet+ &  {36.28/0.9736} & {39.23/0.9872} & {41.81/0.9937} & {{45.03}}/{0.9952} & {45.15}/{0.9959} & {41.50}/{0.9891}\\
		&DICDNet & {37.19}/{0.9853} & {39.53}/{0.9908} & {42.25}/{0.9941} & {44.91}/{0.9953} & {{45.27}}/{0.9958} & {41.83}/{0.9923}\\
		\midrule
		\multirow{3}{*}{\footnotesize{\rotatebox[origin=c]{90}{\makecell[c]{Image-\\domain}} }}
		&MeanFlow   & 28.47/0.8795          & 31.09/0.9253 & 34.92/0.9391 & 37.01/0.9454          & 37.83/0.9489  & 33.86/0.9276\\
		&Refusion   & \textbf{33.66}/0.9694 & 33.82/0.9748 & 39.36/0.9851 & \textbf{43.58}/0.9858 & 40.31/0.9812  & 38.15/0.9793\\
		&Ours       & 32.77/\textbf{0.9725} & \textbf{35.23/0.9796} & \textbf{40.88/0.9860} & 42.10/\textbf{0.9865} & \textbf{42.39/0.9867} & \textbf{38.67/0.9823}\\
		
		\bottomrule
	\end{tabular}
	\label{MAR}
\end{table*}

\begin{table*}[!t]
	\centering
	\small
	\setlength{\tabcolsep}{3pt} 
	
	\caption{Our EDA has superior averaging results for shadow removal, and it gives the best restoration of non-shadow regions.}
	\begin{tabular}{c|ccc|ccc|ccc} 
		\toprule
		\multirow{2}*{Methods} & \multicolumn{3}{c}{Shadow Region (S)} & \multicolumn{3}{c}{Non-Shadow Region (NS)}& \multicolumn{3}{c}{All Image (ALL)}   \\
		
		& PSNR↑ & SSIM↑ & RMSE↓& PSNR↑ & SSIM↑ & RMSE↓& PSNR↑ & SSIM↑ & RMSE↓\\
		\midrule
		Input & 22.40 & 0.936 & 32.10 & 27.32 & 0.976 & 7.09 & 20.56 & 0.893 & 10.88\\
		G2R  & 31.63 & 0.975 & 10.72 & 26.19 & 0.967 & 7.55 & 24.72 & 0.932 & 7.85\\
		DHAN  & 35.53 & 0.988 & 7.49 & 31.05 & 0.971 & 5.30 & 29.11 & 0.954 & 5.66\\
		Fu \emph{et al.} & 34.71 & 0.975 & 7.91 & 28.61 & 0.880 & 5.51 & 27.19 & 0.945 & 5.88\\
		Zhu \emph{et al.}  & 36.95 & 0.987 & 8.29 & 31.54 & \textbf{0.978} & 4.55 & 29.85 & 0.960 & 5.09\\
		BMNet & 35.61  & \textbf{0.988} & 7.60 & 32.80 & 0.976 & 4.59 & 30.28 & 0.959 & 5.02\\ 
		ShadowFormer  &\color{red}\textbf{ 37.99} & \color{red}\textbf{0.990} & \color{red}\textbf{6.16} & \textbf{33.89} & \color{red}\textbf{0.980} & \textbf{3.90} & \textbf{31.81} & \textbf{0.967} & \textbf{4.27}\\
		%
		MeanFlow & 33.13 & 0.979 & 11.10 & 27.23 & 0.910 & 9.50 & 25.77 & 0.880 & 9.77  \\
		Refusion & 34.62 & 0.980 & 8.74 & 28.64 & 0.910 & 6.99 & 27.23 & 0.882 & 7.24\\
		Ours & \textbf{37.80} & \color{red}\textbf{0.990} & \textbf{6.27} & \color{red}\textbf{34.31} & \color{red}\textbf{0.980} & \color{red}\textbf{3.77} & \color{red}\textbf{32.01} & \color{red}\textbf{0.968} & \color{red}\textbf{4.14}\\

		\bottomrule
	\end{tabular}	
	
	\label{SH}
\end{table*}
\endgroup

\subsection{Experimental Setup}
\label{setup}

Our EDA was developed in PyTorch and trained on a single NVIDIA GeForce RTX 3090 GPU.
The parameters are set as, $s(t) = 1$, $\sigma =\sqrt{1-\bar{\alpha_t}}$, where ${\bar{\alpha_t}}$ is hyperparameter as in the DDPM \cite{DDPM}. The total step during training is $T = 100$. 
The training time strategies are denoted as $P_{time}\sim U(0,T)$.
Task-specific training settings (e.g., input resolutions, epochs, and batch sizes) 
are provided in the Supplementary Material E.

Our EDA was compared with task-specific methods. The BFC methods included N4 \cite{N4}, MICO \cite{MICO}, LapGM \cite{LapGM}, ABCNet  \cite{abcnet}. The MAR included LI \cite{kalender1987reduction}, CNNMAR \cite{zhang2018convolutional}, DSCMAR \cite{yu2020deep}, DuDoNet \cite{lin2019dudonet}, InDuDoNet+ \cite{wang2023indudonet+}, DICDNet \cite{wang2021dicdnet}, which are all uses of dual domains (sinusoidal and image domain). 
The SR included G2R \cite{liu2021shadow}, DHAN \cite{cun2020towards}, Fu \emph{et al.} \cite{fu2021auto}, Zhu \emph{et al.} \cite{zhu2022efficient}, BMNet \cite{zhu2022bijective}, ShadowFormer \cite{guo2023shadowformer}.
Furthermore, our experiments involved representative diffusion methods,
including Refusion \cite{luo2023refusion}, MeanFlow \cite{geng2025meanflow}, and an optimized version of IRSDE \cite{IRSDE}.
Refusion is a representative Gaussian diffusion-based restoration method which secured second place overall (first in perceptual metrics) in the CVPR NTIRE Shadow Removal Challenge. 
MeanFlow represents the latest advance in flow-based methods, achieving state-of-the-art (SOTA) performance in one-step image generation. Its flexible diffusion process enables its direct application to image restoration.
All results were either quoted from original publications or reproduced using their official implementations, with Peak Signal-to-Noise Ratio (PSNR) and Structural Similarity (SSIM) as the primary evaluation metrics.

\subsection{Experimental Results} 
\label{result}
The qualitative and quantitative results show that our EDA is more efficient and accurate in restoration. With only \textbf{5 sampling steps}, EDA outperforms the \textbf{100-step} Refusion and most task-specific methods, achieving SOTA performance in BFC and SR.
The MeanFlow training objective fundamentally struggles to align and minimize the complex Task-Specific Error Distribution characteristic of inverse problems. This results in a significant inherent bias in the MeanFlow prediction, leading to sub-optimal performance compared to SOTA restoration methods in the PSNR/SSIM-based precise recovery tasks. Consequently, its scores are excluded from the qualitative analysis tables.

\textbf{Bias Field Correction in MRI.} 
Quantitative (Tab. \ref{BC}) and qualitative results (Fig. \ref{results1}) demonstrate that our EDA effectively removes bias fields. The correlation coefficient (COCO) and coefficient of variation (CV) for both gray matter (GM) and white matter (WM) are additionally reported.
EDA significantly enhances correction accuracy, as evidenced by the highest PSNR, SSIM, and COCO. Additionally, the lowest CV for WM and GM reflects that EDA has the most homogeneous intensity within the tissue, ensuring the homogeneity required for precise subsequent tissue segmentation. 
MeanFlow exhibits the highest CV for GM ($15.49$) among all methods, indicating that its ODE training process struggles to achieve the necessary tissue intensity uniformity.
Notably, the restoration speed of EDA is  $0.182$ sec/slice, significantly faster than the $9.665$ sec/slice of Refusion, representing a substantial $\sim53$-fold acceleration, enabling EDA viable for high-throughput clinical applications. Fig. \ref{steps} shows that EDA requires fewer sampling steps while yielding more accurate results.
Fig. \ref{results1} shows the superior fidelity of our corrected images and the high accuracy of predicted bias fields (second row in Fig. \ref{results1}).

\textbf{Metal Artifact Reduction in CT.} 
Quantitative (Tab. \ref{MAR}) and qualitative results (Fig. \ref{results1}) demonstrate that our EDA effectively reduces metal artifacts, exhibiting competitive performance and accurately restoring tissue structure. 
Notably, all MAR-specific methods use dual-domain (sinusoidal and image domain) information for restoration, whereas our EDA and Refusion rely only on image-domain information.
A comprehensive evaluation was performed across five test data groups stratified by metal size. Using only the image domain, our EDA outperforms several dual-domain methods, showing that our EDA has strong image restoration performance, but a small performance gap remains between EDA and SOTA dual-domain methods. 
MeanFlow exhibits low average performance because the task involves severe data loss and multiple plausible restorations, but ODE diffusion produces an averaged solution, resulting in blurring in artifact regions.
Additionally, the overall results of the EDA with 5-step sampling are better than the 100-step Refusion in Tab. \ref{MAR}, and the sampling pace and performance of EDA notably outperforms Refusion as shown in Fig. \ref{steps}.

\textbf{Shadow Removal in Natural Images.} 
Quantitative (Tab. \ref{SH}) and qualitative (Fig. \ref{results1}) results demonstrate that our EDA removes shadows most accurately and provides the most realistic restoration of non-shadow regions.
The root mean square error (RMSE) in the LAB color space is reported additionally.  
Our method achieves the best performance of the overall image, as measured by PSNR, SSIM, and RMSE, which is also visually confirmed in Fig. \ref{results1}.
The EDA restoration model is based on ShadowFormer, and EDA achieves improved performance, highlighting its effectiveness.
Crucially, EDA maintains the highest fidelity in the Non-Shadow (NS) region (PSNR $34.31$ dB, RMSE $3.77$), demonstrating that the generalized diffusion framework preserves original image quality better than all competitors, indicating EDA's ability to precisely delineate the shadow boundary and avoid introducing artifacts into the background.
Our SDE diffusion framework significantly outperforms ODE MeanFlow on this task. Shadow removal is ill-posed problem that requires the model to generate plausible, high-frequency textures and colors that were occluded by the shadow \cite{qin2025reversingflow}. MeanFlow's ODE training follows an averaged trajectory, which leads to a high ALL RMSE of $9.77$. In contrast, the SDE-based stochasticity integrated by EDA explores the manifold of plausible restorations and corrects errors, which is critical for finding a high-fidelity solution.
Our EDA significantly enhances both the sampling efficiency and the de-shading effect compared to Refusion, as shown in Fig. \ref{steps}.

\FloatBarrier
\section{Conclusion} 
This paper introduces EDA, a unified design space for diffusion models with arbitrary noise patterns for universal image restoration. EDA generalizes vanilla design space to multivariate Gaussian distributions and rigorously demonstrates its ability to diffuse arbitrary noise patterns. Furthermore, our study proves that extending the diffused noise from simple Gaussian patterns, as used in EDM, to more complex arbitrary patterns introduces no additional computational overhead.
EDA has been validated on MRI bias field correction, CT metal artifact reduction, and natural image shadow removal, achieving superior accuracy and efficiency in fewer than $5$ steps as shown in Fig. \ref{steps}.

\textbf{Limitations:}
Compared to deterministic diffusion methods, such as Flow Matching (ODE diffusion) and Cold Diffusion (operator-based), EDA employs an SDE-driven diffusion process where inherent \emph{stochasticity} enhances robustness and performance. While Prop. \ref{prop1} presents three configurations for all potential scenarios, EDM provides a smooth transition that balances \emph{stochasticity} and \emph{flexibility}. 
Case 1 maximizes randomness by theoretically spanning the entire noise domain via weighted basis functions. In contrast, the more widely applicable Cases 2–3 reduce randomness through only stochastic scalar weight adjustments, resembling deterministic methods. Consequently, EDA’s design entails an inherent trade-off between diffusion stochasticity and application scope.

\section{Acknowledgements} 
This work was supported by the National Natural Science Foundation of China under Grant No. 82527807; the Shenzhen Medical Research Fund under Grant No. C2501016; the Science and Technology Innovation Committee of Shenzhen Municipality under Grant No. JCYJ20250604145426036; the Key Research \& Development Program of Heilongjiang Province under Grant Nos. 2023ZX01A08;

{
    \small
    \bibliographystyle{ieeenat_fullname}
    \bibliography{main}
}
\clearpage
\appendix 
\maketitlesupplementary

\section{Derivation of $f(\cdot)$, $g(\cdot)$ and $\phi(\cdot)$ in SDE}
\label{sec:derivation}
\label{A}
\subsection*{Problem Setup}
Given the EDA forward process:
\begin{equation}
	x_{t} = s(t)x_{0} + s(t)\sigma(t)\sum_{m=1}^{M} \frac{\epsilon_{m} + \eta}{\eta + 1} h_{m,\bm{x}_0},
\end{equation}
and its distribution:
\begin{equation}
    \label{21}
    \begin{aligned}
    P(x_t \mid x_0)
      &= \mathcal{N}\Bigl(
          x_t;\,
          s(t)x_0
          + \frac{\eta\, s(t)\, \sigma(t)}{\eta + 1}
            \sum_{m=1}^{M} h_{m,\bm{x}_0},\\[-1mm]
      &\hspace{6em}
          \frac{s^{2}(t)\, \sigma^{2}(t)}{(\eta + 1)^{2}}
          \Sigma_{x_0}
         \Bigr).
    \end{aligned}
    \end{equation}
where \(\Sigma_{x_{0}} = H_{x_{0}}H_{x_{0}}^{T}\) and \(H_{x_{0}} = \begin{bmatrix} h_{1,\bm{x}_0} & \ldots & h_{M,\bm{x}_0} \end{bmatrix}\). The associated Stochastic Differential Equation (SDE) is expressed as:
\begin{equation}
	dx = \left[ f(t)x + \phi(t, x_{0}) \right] dt + g(t) \sum_{m=1}^{M} h_{m,\bm{x}_0} dw_{t}^{(m)},
\end{equation}
where \(dw_{t}^{(m)}\) are independent Wiener processes satisfying \(\mathbb{E}[dw_{t}^{(m)} dw_{t}^{(n)}] = \delta_{mn} dt\). The coefficient functions are derived as follows:

\begin{align}
	f(t) &= \frac{s'(t)}{s(t)} \label{eq:f} \\
	\phi_{x_{0}}(t) &= \frac{\eta s(t)\sigma'(t)}{\eta + 1} \sum_{m=1}^{M}  h_{m,\bm{x}_0} \label{eq:phi} \\
	g(t) &= \frac{s(t)}{\eta + 1} \sqrt{\frac{d\sigma^{2}(t)}{dt}} \label{eq:g}
\end{align}

\textbf{Derivation:}
The derivation is performed in two parts: Mean and Variance.

\subsubsection*{1. Derivation of the Mean Component}

The mean of the forward diffusion process $P(x_t | x_0)$ is given by:
\begin{equation}
	\mu(t) = s(t)x_{0} + \frac{\eta s(t)\sigma(t)}{\eta + 1}\sum_{m=1}^{M} h_{m,\bm{x}_0}
\end{equation}

For the SDE:
\begin{equation}
	\mathrm{d}x = \left[f(t)x + \phi_{x_{0}}(t)\right]\mathrm{d}t + g(t) \sum_{m=1}^{M} h_{m,\bm{x}_0} \mathrm{d}w_t^{(m)}
\end{equation}
the differential equation governing the mean $\mu(t)$ is:
\begin{equation}
	\frac{\mathrm{d}\mu(t)}{\mathrm{d}t} = f(t)\mu(t) + \phi_{x_{0}}(t), \quad \mu(0) = x_0
\end{equation}

Solving this first order linear non-homogeneous differential equation by the integrating factor method with $\lambda(t)=\exp\left(-\int_0^t f(\tau)\mathrm{d}\tau\right)$ yields:

\begingroup\small
\begin{align*}
&\mu(t)
= \exp\!\Bigl(\!\int_0^t f(\tau)\,\mathrm{d}\tau\!\Bigr) x_0 \notag\\
& + \exp\!\Bigl(\!\int_0^t f(\tau)\,\mathrm{d}\tau\!\Bigr)
   \int_0^t \phi(\tau)
   \exp\!\Bigl(\!-\!\int_0^\tau f(s)\,\mathrm{d}s\!\Bigr)\,\mathrm{d}\tau.
\end{align*}
\endgroup
    
Equating this to the mean of $P(x_t | x_0)$, we identify $s(t) = \exp\left( \int_0^t f(\tau)\mathrm{d}\tau \right)$. So we obtain:
\begin{equation}
	f(t) = \frac{s'(t)}{s(t)} \tag*{\qed}
\end{equation}

And for the second term:
\begin{equation}
	s(t) \int_0^t \frac{\phi_{x_{0}}(\tau)}{s(\tau)} \mathrm{d}\tau = \frac{\eta s(t)\sigma(t)}{\eta + 1}\sum_{m=1}^{M} h_{m,\bm{x}_0}
\end{equation}

Canceling $s(t)$ from both sides and differentiating with respect to $t$:
\begin{equation}
	\frac{\phi_{x_{0}}(t)}{s(t)} = \frac{\mathrm{d}}{\mathrm{d}t}\left[\frac{\eta \sigma(t)}{\eta + 1}\sum_{m=1}^{M} h_{m,\bm{x}_0}\right]
\end{equation}

Because $\sum_{m=1}^M h_{m,\bm{x}_0}$ is time-independent, we finally obtain:
\begin{align}
	\phi_{x_{0}}(t) &=\frac{\eta s(t) \sigma'(t)}{\eta + 1}\sum_{m=1}^{M} h_{m,\bm{x}_0} \tag*{\qed}
\end{align}

\subsection*{2. Derivation of the Variance Component}

The covariance matrix of the forward diffusion process is given by:
\begin{equation}
	\Sigma(t) = \frac{s^{2}(t)\sigma^{2}(t)}{(\eta + 1)^{2}} \Sigma_{x_{0}}, \quad 
\end{equation}
where $ \Sigma_{x_{0}} = \sum_{m=1}^M h_{m,\bm{x}_0} h_{m,\bm{x}_0}^\top$.

Applying Itô's lemma to the covariance matrix $\Sigma_{S}(t)$ of the SDE, the differential equation is satisfied:
\begin{equation}
	\small\frac{\mathrm{d}\Sigma_{S}(t)}{\mathrm{d}t} = 2f(t)\Sigma_{S}(t) + g^2(t) \sum_{m=1}^M h_{m,\bm{x}_0} h_{m,\bm{x}_0}^\top, \quad \Sigma(0) = 0
\end{equation}

Solving this equation, and we obtain:
\begin{align}
	\Sigma_{S}(t) = s^2(t) \int_0^t \frac{g^2(\tau)}{s^2(\tau)} \mathrm{d}\tau \cdot \sum_{m=1}^M h_{m,\bm{x}_0} h_{m,\bm{x}_0}^\top
\end{align}

Equating $\Sigma_{S}(t)$ to the forward process covariance $\Sigma(t)$:
\begin{equation}
	\frac{s^{2}(t)\sigma^{2}(t)}{(\eta + 1)^{2}} \Sigma_{x_{0}} = s^2(t) \int_0^t \frac{g^2(\tau)}{s^2(\tau)} \mathrm{d}\tau \cdot \Sigma_{x_{0}}
\end{equation}
Canceling $s^2(t)\Sigma_{x_{0}}$ from both sides yields:
\begin{equation}
	\int_0^t \frac{g^2(\tau)}{s^2(\tau)} \mathrm{d}\tau = \frac{\sigma^{2}(t)}{(\eta + 1)^{2}}
\end{equation}

Differentiating both sides with respect to $t$:
\begin{equation}
	\frac{g^2(t)}{s^2(t)} = \frac{1}{(\eta + 1)^{2}} \frac{\mathrm{d}\sigma^2(t)}{\mathrm{d}t}
\end{equation}
Taking the square root and solving for $g(t)$, we obtain:
\begin{equation}
	g(t) = \frac{s(t)}{(\eta + 1)} \sqrt{\frac{\mathrm{d}\sigma^2(t)}{\mathrm{d}t}} \tag*{\qed}
\end{equation}

\section{Derivation of Probability Flow ODE for EDA}
\label{B}
\subsection*{Problem Setup}
Consider the forward process SDE for EDA:
\begin{equation}
	\mathrm{d}x = \left[f(t)x + \phi_{x_{0}}(t)\right]\mathrm{d}t + g(t) \sum_{m=1}^{M} h_{m,\bm{x}_0} \mathrm{d}w_t^{(m)}
\end{equation}
where:
\begin{itemize}
	\item $f(t)$: Scalar drift coefficient
	\item $\phi_{x_{0}}(t)$: Deterministic offset term
	\item $g(t)$: Diffusion coefficient
	\item $h_{m,\bm{x}_0}$: Basis functions (only dependent on initial condition $x_0$)
	\item $\mathrm{d}w_t^{(m)}$: Independent Wiener process increments
\end{itemize}

Our goal is to find the corresponding Probability Flow ODE (PFODE):
\begin{equation}
	\frac{\mathrm{d}x}{\mathrm{d}t} = v_{x_0}(x, t)
\end{equation}
whose solution preserves the probability distribution $p(x|x_0)$ of the original SDE.

\subsection*{Derivation via Fokker-Planck Framework}
For a general SDE $\mathrm{d}x_t = \mu(x_t,t)\mathrm{d}t + \bar{\Sigma}(x_t,t)\mathrm{d}w_t$, the Fokker-Planck equation (FPE) is:
\begin{equation}
	\frac{\partial p}{\partial t} = -\nabla \cdot \big[\mu p\big] + \frac{1}{2} \nabla \nabla : \big[\bar{\Sigma} \bar{\Sigma}^\top p\big]
\end{equation}
where $\nabla \cdot$ denotes divergence and $\nabla \nabla :$ denotes the double divergence operator.

Substituting our SDE coefficients:
\begin{itemize}
	\item Drift term: $\mu(x,t) = f(t)x + \phi_{x_{0}}(t)$
	\item Diffusion term: $\bar{\Sigma}(x,t) = g(t) \cdot [h_{1,\bm{x}_0}, \dots, h_{M,\bm{x}_0}]$
\end{itemize}
The covariance matrix becomes:
\begin{equation}
	\bar{\Sigma} \bar{\Sigma}^\top = g^2(t) \sum_{m=1}^M h_{m,\bm{x}_0} h_{m,\bm{x}_0}^\top \triangleq g^2(t) \Sigma_{x_{0}}
\end{equation}

The FPE specializes to:
\begin{equation}
	\frac{\partial p}{\partial t} = -\nabla \cdot \big[(f(t)x + \phi_{x_{0}}(t))p\big] + \frac{1}{2} g^2(t) \nabla \nabla : \big[\Sigma_{x_{0}} p\big]
\end{equation}

\subsection*{Constructing the PFODE}
The probability flow ODE requires the continuity equation:
\begin{equation}
	\frac{\partial p}{\partial t} + \nabla \cdot \big[v_{x_0}(x, t)p\big] = 0
\end{equation}

Equating with the FPE, we obtain:
\begin{equation}
	\nabla \cdot \big[v_{x_0} p\big] = \nabla \cdot \big[(f(t)x + \phi_{x_{0}}(t))p\big] - \frac{1}{2} g^2(t) \nabla \nabla : \big[\Sigma p\big]
\end{equation}

\subsubsection*{Solving for the Velocity Field}
Introduce the \textbf{score function}:
\begin{equation}
	\bar{s}(x,t) = \nabla_x \ln p(x|x_0) = \frac{\nabla_x p}{p}
\end{equation}
For the diffusion term:
\begin{align*}
	\nabla \nabla : \big[\Sigma_{x_{0}} p\big] &= \sum_{i,j} \frac{\partial^2}{\partial x_i \partial x_j} \big[ (\Sigma_{x_{0}})_{ij} p \big] \\
	&= \nabla \cdot \left[ \Sigma_{x_{0}} \nabla p \right] \\
	&= \nabla \cdot \left[ \Sigma_{x_{0}} p \nabla \ln p \right] \quad \text{(using } \nabla p = p \nabla \ln p \text{)}
\end{align*}
This establishes the critical connection between the diffusion term and the score function. Then we transform the diffusion term:
\begin{align}
	\frac{1}{2} \nabla \nabla : \big[\Sigma_{x_{0}} p\big] &= \nabla \cdot \left[ \frac{1}{2} \Sigma_{x_{0}} \nabla p \right] \\
	&= \nabla \cdot \left[ \frac{1}{2} \Sigma_{x_{0}} p s \right]
\end{align}

Substituting back, we obtain:
\begin{equation}
	\nabla \cdot \big[v_{x_0} p\big] = \nabla \cdot \left[ (f(t)x + \phi_{x_{0}}(t))p - \frac{1}{2} g^2(t) \Sigma_{x_{0}} p \bar{s} \right]
\end{equation}

This identifies the velocity field:
\begin{equation}
	v_{x_0}(x,t) = f(t)x + \phi_{x_{0}}(t) - \frac{1}{2} g^2(t) \Sigma_{x_{0}} \bar{s}(x,t)
\end{equation}

\subsection*{Final PFODE Form}
Substituting $s(x,t) = \nabla_x \ln p(x,t)$, we obtain the complete PFODE:
\begin{equation}
	\frac{\mathrm{d}x}{\mathrm{d}t} = f(t)x + \phi_{x_{0}}(t) - \frac{1}{2} g^2(t) \Sigma_{x_{0}} \nabla_x \ln p(x|x_0) \tag*{\qed}
\end{equation}
where $\Sigma_{x_{0}} = \sum_{m=1}^M h_{m,\bm{x}_0} h_{m,\bm{x}_0}^\top$.

\section{Derivation of Deterministic Sampling Formula from PFODE}
\label{C}
\subsection*{PFODE Simplification}
Given the Probability Flow ODE (PFODE):
\begin{equation}
	\label{PFODE}
	\frac{\mathrm{d}x}{\mathrm{d}t} = f(t)x + \phi_{x_{0}}(t) - \frac{1}{2} g^2(t) \Sigma_{x_{0}} \nabla_x \ln p(x|x_0)
\end{equation}

Substitute the parameter relationships derived in Appendix \ref{A}:
\begin{align}
    f(t)
      &= \frac{s'(t)}{s(t)}, \quad
       \phi_{x_0}(t)
       = \frac{\eta\, s(t)\, \sigma'(t)}{\eta+1}
         \sum_{m=1}^{M} h_{m,\bm{x}_0}, \notag\\
    g^2(t)
      &= \frac{2\, s^2(t)\, \sigma(t)\, \sigma'(t)}{(\eta+1)^2}.
\end{align}

This transforms the PFODE into:
\begin{equation}
    \begin{aligned}
        \frac{\mathrm{d}x}{\mathrm{d}t}
          &= \frac{s'(t)}{s(t)}\,x
           + \frac{\eta\, s(t)\, \sigma'(t)}{\eta + 1}
             \sum_{m=1}^{M} h_{m,\bm{x}_0} \\
          &\quad
           - \frac{s^2(t)\, \sigma(t)\, \sigma'(t)}{(\eta + 1)^2}
             \Sigma_{x_{0}}\,
             \nabla_{x_t}\! \log p(x \mid x_0).
        \end{aligned}
\end{equation}

\subsection*{Denoiser Learning}
Define the loss function $\mathcal{L}_{x_0}(D;\sigma(t))$:
\begin{equation}
	\mathcal{L}_{x_0}(D;\sigma(t)) = \mathbb{E}_{x\sim \mathcal{N}(\mu(x_0,t),\bar{\Sigma}(t))} \|D(x;\sigma) - x_0\|^2
\end{equation}
where:
\begin{align*}
	\mu(x_0,t) &= s(t)x_0 + \frac{s(t)\sigma(t)}{\eta+1}\sum_{m=1}^M \eta h_{m,\bm{x}_0} \\
	\bar{\Sigma}(t) &= \frac{s^2(t)\sigma^2(t)}{(\eta+1)^2}\Sigma_{x_0}
\end{align*}

Expanding the $\mathcal{L}_{x_0}(D;\sigma(t))$ yields:
\begin{align}
    & \mathcal{L}_{x_0}(D;\sigma(t))
      = \mathbb{E}_{x\sim\mathcal{N}(\mu(x_0,t),\bar{\Sigma}(t))}
        \|D(x;\sigma)-x_0\|^2, \\
    & = {\small \int_{\mathbb{R}^d}
        \mathcal{N}\!\Bigl(
          s(t)x_0 + \tfrac{\eta\,s(t)\,\sigma(t)}{\eta+1}\!\sum_{m=1}^M h_{m,\bm{x}_0},\;
          \tfrac{s^2(t)\,\sigma^2(t)}{(\eta+1)^2}\,\Sigma_{x_0}
        \Bigr)\,}\notag\\
    &   \qquad {}\times\|D_{x_0}(x;\sigma)-x_0\|^2\,\mathrm{d}x, \\
    & \triangleq \int_{\mathbb{R}^d}
        \mathcal{L}_{x_0}(D;\sigma(t),x)\,\mathrm{d}x. 
    \end{align}

Then minimize $\mathcal{L}_{x_0}(D;\sigma(t),x)$.
This convex optimization problem is solved by setting the gradient to zero:
\begin{align}
0
  &= \nabla_{D_{x_0}(x,\sigma)}
     \mathcal{L}_{x_0}(D;\sigma(t),x),\\
0
&= \begingroup \small
\begin{aligned}[t]
\nabla_{D_{x_0}(x,\sigma)}\!
\Bigl[
    \mathcal{N}\!\Bigl(
    s(t)x_0
    + \tfrac{s(t)\sigma(t)}{\eta + 1}
        \sum_{m=1}^{M}\eta h_{m,\bm{x}_0},\;\\
    \tfrac{s^2(t)\sigma^2(t)}{(\eta + 1)^2}\Sigma_{x_0}
    \Bigr)\qquad {}\times \|D_{x_0}(x;\sigma)-x_0\|^2
\Bigr]
\end{aligned}
\endgroup\\
0
  &= \nabla_{D_{x_0}(x,\sigma)}
     \|D_{x_0}(x;\sigma)-x_0\|^2.
\end{align}

This leads to the optimal solution:
\begin{equation}
D_{x_0}(x;\sigma) = x_0.
\end{equation}

\subsection*{Score Function Derivation}
For the conditional distribution, we define the mean $\boldsymbol{\mu}_c$ and covariance $\boldsymbol{\Sigma}_c$ as:
\begin{equation}
    \boldsymbol{\mu}_c = s(t)x_0 + \frac{\eta s(t)\sigma(t)}{\eta+1}\sum_{m=1}^M h_{m,\bm{x}_0}, \quad 
    \boldsymbol{\Sigma}_c = \frac{s^2(t)\sigma^2(t)}{(\eta+1)^2}\Sigma_{x_{0}}.
\end{equation}
The distribution is then expressed compactly as $p(x_t|x_0) = \mathcal{N}(x_t; \boldsymbol{\mu}_c, \boldsymbol{\Sigma}_c)$. The score function is derived as:
\begin{align}
    \nabla_{x_t} \log p(x_t|x_0) &= \frac{\nabla_{x_t}\mathcal{N}(x_t; \boldsymbol{\mu}_c, \boldsymbol{\Sigma}_c)}{\mathcal{N}(x_t; \boldsymbol{\mu}_c, \boldsymbol{\Sigma}_c)} \\
    &= \frac{\mathcal{N}(x_t; \boldsymbol{\mu}_c, \boldsymbol{\Sigma}_c) \left[ \boldsymbol{\Sigma}_c^{-1}(\boldsymbol{\mu}_c - x_t) \right]}{\mathcal{N}(x_t; \boldsymbol{\mu}_c, \boldsymbol{\Sigma}_c)} \label{eq:deriv_step}\\
    &= \boldsymbol{\Sigma}_c^{-1}(\boldsymbol{\mu}_c - x_t).
\end{align}
Substituting the definitions of $\boldsymbol{\mu}_c$ and $\boldsymbol{\Sigma}_c$ back yields the final analytical form.

Substituting into the score function yields:
\begin{align}
    \nabla_{x_t} \log p(x_t|x_0)
      &= \frac{(\eta+1)^2}{s^2(t)\sigma^2(t)}\Sigma_{x_0}^{-1}
         \Bigl(
           s(t)x_0 \notag\\
      &    + \frac{\eta\, s(t)\, \sigma(t)}{\eta+1}
             \sum_{m=1}^{M} h_{m,\bm{x}_0}
           - x_t
         \Bigr).
    \end{align}

\subsection*{ODE Simplification}
Substituting the score function into the PFODE:
\begin{align}
	\frac{\mathrm{d}x}{\mathrm{d}t} &= \frac{s'(t)}{s(t)}x + \frac{\eta s(t)\sigma'(t)}{\eta+1}\sum_{m=1}^M h_{m,\bm{x}_0} \notag\\
    &- \frac{s^2(t)\sigma(t)\sigma'(t)}{(\eta+1)^2} \Sigma_{x_{0}} \left(\frac{(\eta+1)^2}{s^2(t)\sigma^2(t)}\Sigma_{x_{0}}^{-1}\right. \nonumber\\
	& \quad \left.\left(s(t)x_0 + \frac{\eta s(t)\sigma(t)}{\eta+1}\sum_{m=1}^M  h_{m,\bm{x}_0} - x\right)\right) \label{69}\\
	&= \frac{s'(t)}{s(t)}x + \frac{\eta s(t)\sigma'(t)}{\eta+1}\sum_{m=1}^M h_{m,\bm{x}_0} \notag\\
    &- \frac{\sigma'(t)}{\sigma(t)} \left(s(t)x_0 + \frac{\eta s(t)\sigma(t)}{\eta+1}\sum_{m=1}^M  h_{m,\bm{x}_0} - x\right)\\
	&= \left(\frac{s'(t)}{s(t)} + \frac{\sigma'(t)}{\sigma(t)}\right)x_t - \frac{\sigma'(t)s(t)}{\sigma(t)}x_0\\
	&= \left(\frac{s'(t)}{s(t)} + \frac{\sigma'(t)}{\sigma(t)}\right)x_t - \frac{\sigma'(t)s(t)}{\sigma(t)}D_{x_0}(x;\sigma)\\
\end{align}

Use a uniform network $D(x;\sigma)$ to learn all $D_{x_0}(x;\sigma)$, the loss function is:
\begin{align}
	&\mathcal{L}(D;\sigma(t)) \notag\\
    &= \mathbb{E}_{y\sim P_{\text{data}}}\mathbb{E}_{x\sim \mathcal{N}(\mu(y,t),\bar{\Sigma}(t))} \|D(x;\sigma) - D_{y}(x;\sigma)\|^2\\
	&= \mathbb{E}_{y\sim P_{\text{data}}}\mathbb{E}_{x\sim \mathcal{N}(\mu(y,t),\bar{\Sigma}(t))} \|D(x;\sigma) - y\|^2\\
\end{align}
when the network is fitted:
\begin{align}
	\label{78}
	D(x;\sigma) \approx D_{y}(x;\sigma)=y, \quad \forall y\sim P_{\text{data}}
\end{align}
Note that the $\approx$ here has an error and is not the theoretical optimal point of a convex problem. When the error is small enough during training, substituting the $D(x;\sigma)$ into the simplified PFODE:
\begin{align}
	\label{samp} 
	\frac{\mathrm{d}x}{\mathrm{d}t} = \left(\frac{s'(t)}{s(t)} + \frac{\sigma'(t)}{\sigma(t)}\right)x - \frac{\sigma'(t)s(t)}{\sigma(t)}D(x;\sigma) \tag*{\qed} 
\end{align}

Notably, all terms in the derivation of Eq. \ref{69} related to the data-dependent basis $h_{m,\bm{x}_0}$ simplify entirely. 
This results in the deterministic sampling procedure for our any-noise-based EDA becoming identical to that of the EDM framework Eq. \ref{77}.
\section{Derivation of Optimal Case for EDA}
\label{D}
\begin{Proposition}
	The approximation in Eq. \ref{78} of Appendix \ref{C} introduces errors by substituting the denoiser output $D(x;\sigma)$ with the ground-truth training target $y$ upon training convergence.  However, these errors vanish when the basis set is independent of data sample $x_0$, $H = [h_1,\ldots,h_M]$, in contrast to the sample-dependent basis \(H_{x_{0}} = \begin{bmatrix} h_{1,\bm{x}_0},..., h_{M,\bm{x}_0} \end{bmatrix}\) in Eq. \ref{21}. This represents the optimal case for EDA.
\end{Proposition}

\emph{Proof.}
When the basis set $H$ is independent of data sample $x_0$, the PFODE Eq. \ref{PFODE} could use the marginal distribution $p(x)$ to substitute the conditional $p(x|x_0)$, which more accurately represents the distribution of $x_t$. The covariance matrix becomes:
\begin{equation}
	\mathbf{\Sigma} = \sum_{m=1}^M h_m h_m^\top
\end{equation}
yielding the modified PFODE:
\begin{equation}
	\label{79}
	\frac{\mathrm{d}x}{\mathrm{d}t} = f(t)x + \phi(t) - \frac{1}{2}g^2(t)\mathbf{\Sigma}\nabla_x \log p(x)
\end{equation}
with coefficients:
\begin{align}
	f(t) = \frac{s'(t)}{s(t)}, \quad \phi(t) = \frac{\eta s(t)\sigma'(t)}{\eta+1}\sum_{m} h_m, \notag\\
    \quad g(t) = \frac{s(t)}{\eta+1}\sqrt{\frac{\mathrm{d}\sigma^2(t)}{\mathrm{d}t}}
\end{align}

Suppose our training set consists of a finite samples $[y_1,\ldots,y_Y]$. Thus $p_{\text{data}}(x)$ is represented by a mixture of Dirac delta distributions:
\begin{equation}
	p_{\text{data}}(x) = \frac{1}{Y}\sum_{i=1}^Y \delta(x - y_i)
\end{equation}
The marginal distribution $p(x)$ is:
\begin{align}
    p(x)
    &= \int_{\mathbb{R}^d} p_{\text{data}}(x_0)\,
       \mathcal{N}\!\bigl(x;\,\mu(x_0,t),\,\Sigma_t\bigr)\,\mathrm{d}x_0 \\
    &= \int_{\mathbb{R}^d} \tfrac{1}{Y}\!\sum_{i=1}^Y \delta(x_0-y_i)\,
       \mathcal{N}\!\bigl(x;\,\mu(x_0,t),\,\Sigma_t\bigr)\,\mathrm{d}x_0 \\
    &= \tfrac{1}{Y}\!\sum_{i=1}^Y \int_{\mathbb{R}^d} \delta(x_0-y_i)\,
       \mathcal{N}\!\bigl(x;\,\mu(x_0,t),\,\Sigma_t\bigr)\,\mathrm{d}x_0 \\
    &= \tfrac{1}{Y}\!\sum_{i=1}^Y
       \mathcal{N}\!\bigl(x;\,\mu(y_i,t),\,\Sigma_t\bigr),
    \label{86}
    \end{align}
    \vspace{-0.5ex}
    where
    \begin{align}
    \mu(x_0,t)
    &= s(t)x_0 + \tfrac{\eta\, s(t)\, \sigma(t)}{\eta+1}\!\sum_m h_m, \qquad \notag\\
    &\Sigma_t = \tfrac{s^2(t)\, \sigma^2(t)}{(\eta+1)^2}\boldsymbol{\Sigma}.
    \end{align}

Now consider the training objective:
\begingroup\small
\begin{align}
    \mathcal{L}(D;\sigma)
      &= \mathbb{E}_{y\sim p_{\text{data}}}
         \mathbb{E}_{x\sim p(x|y)}\|D(x;\sigma)-y\|^2,\\
      &= \mathbb{E}_{y\sim p_{\text{data}}}
         \int_{\mathbb{R}^d}
         \mathcal{N}\!\bigl(x;\mu_y(t),\bar{\Sigma}(t)\bigr)
         \|D(x;\sigma)-y\|^2\,\mathrm{d}x,\\
      &= \tfrac{1}{Y}\!\sum_{i=1}^Y
         \int_{\mathbb{R}^d}
         \mathcal{N}\!\bigl(x;\mu_{y_i}(t),\bar{\Sigma}(t)\bigr)
         \|D(x;\sigma)-y_i\|^2\,\mathrm{d}x,\\
      &= \int_{\mathbb{R}^d}\!
         \tfrac{1}{Y}\!\sum_{i=1}^Y
         \mathcal{N}\!\bigl(x;\mu_{y_i}(t),\bar{\Sigma}(t)\bigr)
         \|D(x;\sigma)-y_i\|^2\,\mathrm{d}x
         \label{90}\\
      &\triangleq
         \int_{\mathbb{R}^d}
         \mathcal{L}(D;x,\sigma)\,\mathrm{d}x.
\end{align}
\endgroup
where
\[
\begin{aligned}
\mu_y(t)
  &= s(t)y + \frac{\eta\, s(t)\, \sigma(t)}{\eta+1}\sum_m h_m, \\
\bar{\Sigma}(t)
  &= \frac{s^2(t)\sigma^2(t)}{(\eta+1)^2}\mathbf{\Sigma}.
\end{aligned}
\]

We minimize $\mathcal{L}(D;\sigma)$ by pointwise minimization: for each fixed $x$, the objective reduces to a weighted least-squares problem
\begin{align}
0
  &= \nabla_{D(x;\sigma)} \mathcal{L}(D;x,\sigma) \\
  &= \nabla_{D(x;\sigma)} \tfrac{1}{Y}\sum_{i=1}^Y w_i(x)\,\|D(x;\sigma)-y_i\|^2 \\
  &= \tfrac{2}{Y}\sum_{i=1}^Y w_i(x)\,\bigl(D(x;\sigma)-y_i\bigr), \\
\Rightarrow\quad
D(x;\sigma)
  &= \frac{\sum_{i=1}^Y w_i(x)\,y_i}{\sum_{i=1}^Y w_i(x)}. \label{95}
\end{align}

where
\[
\begin{aligned}
w_i(x) &:= \mathcal{N}\!\bigl(x;\mu_{y_i}(t),\bar{\Sigma}(t)\bigr),\\
\mu_{y_i}(t) &= s(t)y_i+\tfrac{\eta\,s(t)\,\sigma(t)}{\eta+1}\sum_m h_m,\\
\bar{\Sigma}(t) &= \tfrac{s^2(t)\sigma^2(t)}{(\eta+1)^2}\mathbf{\Sigma}.
\end{aligned}
\]
and $D(x;\sigma)$ denotes the pointwise minimizer (denoiser output) at location $x$ and noise level $\sigma$.

Substituting the marginal distribution  $p(x)$ in Eq. \ref{86}, the score function becomes:
\begin{align}
    \nabla_{x_t}\log p(x_t)
      &= \frac{\nabla_{x_t}p(x_t)}{p(x_t)} \\[2pt]
      &= \frac{
          \nabla_{x_t}\tfrac{1}{Y}\!\sum_{i=1}^Y
          \mathcal{N}\!\bigl(x_t;\mu_i(t),\bar{\Sigma}(t)\bigr)
         }{
          \tfrac{1}{Y}\!\sum_{i=1}^Y
          \mathcal{N}\!\bigl(x_t;\mu_i(t),\bar{\Sigma}(t)\bigr)
         } \\[2pt]
      &= \frac{
          \sum_{i=1}^Y
          \nabla_{x_t}\mathcal{N}\!\bigl(x_t;\mu_i(t),\bar{\Sigma}(t)\bigr)
         }{
          \sum_{i=1}^Y
          \mathcal{N}\!\bigl(x_t;\mu_i(t),\bar{\Sigma}(t)\bigr)
         }.
         \label{99}
    \end{align}
    
    where the term $\nabla_{x_t}\mathcal{N}(\cdot)$ is given by:
    \begin{align}
    &\nabla_{x_t}\mathcal{N}\!\bigl(x_t;\mu_i(t),\bar{\Sigma}(t)\bigr)
      = \mathcal{N}\!\bigl(x_t;\mu_i(t),\bar{\Sigma}(t)\bigr) \notag\\
      &  \nabla_{x_t}\!\Bigl[
           -\tfrac{1}{2}
           (x_t-\mu_i(t))^{\!\top}\!
           \bar{\Sigma}^{-1}(x_t-\mu_i(t))
         \Bigr]\\[2pt]
      &= \mathcal{N}\!\bigl(x_t;\mu_i(t),\bar{\Sigma}(t)\bigr)\,
         \bar{\Sigma}^{-1}\!
         \bigl(\mu_i(t)-x_t\bigr)\notag\\
      &   \cdot\tfrac{(\eta+1)^2}{s^2(t)\sigma^2(t)}.
    \end{align}

    Substituting $\nabla_{x_t}\mathcal{N}(\cdot)$ into Eq.~\ref{99} yields:
\begin{align}
\nabla_{x_t}\log p(x_t)
  &= A(t)
     \frac{\sum_i \mathcal{N}_i\,[s(t)y_i + B(t) - x_t]}
          {\sum_i \mathcal{N}_i} \notag\\
  &= A(t)\!\left[
      s(t)\frac{\sum_i \mathcal{N}_i y_i}{\sum_i \mathcal{N}_i}
      + B(t) - x_t
     \right]. \label{100}
\end{align}

Substituting $D(x_t;\sigma)$ gives:
\begin{align}
\nabla_{x_t}\log p(x_t)
  &= A(t)\,[s(t)D(x_t;\sigma) + B(t) - x_t].
   \label{101}
\end{align}

Finally, substituting Eq.~\ref{101} into the PFODE yields the deterministic sampling form:
\begin{align}
\frac{\mathrm{d}x}{\mathrm{d}t}
 &= \frac{s'(t)}{s(t)}x
   + \frac{\eta s(t)\sigma'(t)}{\eta+1}\!\sum_m h_m\notag\\
 & - s^2(t)\sigma(t)\sigma'(t)\! \left[\frac{A(t)}{(\eta+1)^2}
     (s(t)D(x;\sigma)+B(t)-x)\right] \notag\\
 &= \!\left(\frac{s'(t)}{s(t)}+\frac{\sigma'(t)}{\sigma(t)}\right)x
   -\frac{\sigma'(t)s(t)}{\sigma(t)}D(x;\sigma).
   \tag*{\qed}
   \label{102}
\end{align}

\noindent
where for compactness we define:
\[
\begin{aligned}
\mathcal{N}_i
 &:= \mathcal{N}\!\Bigl(x_t;\,
    s(t)y_i+\tfrac{\eta s(t)\sigma(t)}{\eta+1}\!\sum_m h_m,\,
    \tfrac{s^2(t)\sigma^2(t)}{(\eta+1)^2}\mathbf{\Sigma}\Bigr),\\
A(t)
 &:= \tfrac{(\eta+1)^2}{s^2(t)\sigma^2(t)}\mathbf{\Sigma}^{-1},\qquad
B(t)
 := \tfrac{\eta s(t)\sigma(t)}{\eta+1}\!\sum_{m=1}^M h_m.
\end{aligned}
\]

To summarize, there are two distinctions exist between this framework and Appendix \ref{C}: (1) The marginal probability density Eq. \ref{79} replaces the conditional density Eq. \ref{PFODE} in PFODE, for more accurate representation of $x_t$ distribution; 
and (2) As shown in Eq. \ref{95}, the derivation of the optimal output of denoiser $D(x;\sigma)$ employs a rigorous convex optimization, rather than the approximation with errors as in Eq. \ref{78}. 
Consequently, when the basis set $H$ is independent of the data sampling $x_0$, the theoretical framework achieves greater completeness and represents the optimal case for EDM.

\section{Implementation Details}
\label{E}
\subsection{Specific Experimental Setup}
\label{E1}
For BFC, the mediator $\eta = 0$, the basis set $H$ was set as \textbf{Case 1} in Prop. \ref{prop1}. The input MRIs were $256 \times 256$ and trained for $500$ epochs with a batch size of $1$. For MAR and SR, the $\eta = 10$ and the basis set $H$ followed \textbf{Case 2} in Prop. \ref{prop1}. The input CTs were $416 \times 416$ and trained $1000$ epochs with a batch size of $1$. The Shadow Images were $256 \times 256\times 3$, trained $2000$ epochs with a batch size of $4$. 

\subsection{Bias Field correction in MRI}
\label{E2}
We use the Adam with the momentum as $(0.9,0.999)$. The initial learning rate is set to $2\times 10^{-5}$, decay to $1e^{-8}$ with factor is $0.6$ and the patient is $25$ epochs. We use $0.9999$ Exponential Moving Average (EMA). GDMP is trained for $800$ epochs and the batch size is $1$.
The total diffusion steps $T$ are set to $100$. The noise schedule is defined as $\bar{\alpha}=\prod_{s=1}^t \alpha_s$ as in \cite{DDPM,DDIM}, where $\alpha_t=1-\beta_t$ and $\beta_t$ is linearly increasing from $0.0001$ to $0.02$.

Since the relationship between bias fields and MRI is usually modeled as a multiplicative, we first apply logarithmic transformation to the image, 
converting the multiplicative relationship between the image and noise into an additive one, as $log(A\times B)=log(A)+log(B)$. The $\eta$ is set to zero. 
If the diffused noise is set to be bias field, its smoothness properties must to be satisfied, so we set the basis set $H$ in Eq. \ref{6} to include low-order Legendre polynomials and slowly varying trigonometric functions.
We denote the i-th Legendre polynomial by $P_i(x)$. Then the two-dimensional Legendre polynomial is $P_{m,n}(x,y)=P_m(x)P_n(y)$. The two-dimensional Legendre polynomials with the highest degree less than or equal to $N$ are used as basis functions, $L_N(x,y)\triangleq\{P_{m,n}(x,y)|m+n\le N\}$, where $N$ is a hyperparameter.
We use a rotation function $f(x,y,\theta)=x cos(\theta)+y sin(\theta)$ and define $S_n(x,y) = \{cos(n \cdot f(x,y,\theta)),sin(n \cdot f(x,y,\theta))|mod(\theta,10)=0, \theta\le180\}$. Trigonometric basis set is $T_N(x,y)\triangleq\{S_n(x,y)|2\le n \le N\}$, where $N$ is a hyperparameter. The functions with $n$ less than $2$ are removed because they change too slowly.
In summary, the basis set is:
\begin{equation}
	H_{N_1,N_2}(x,y) \triangleq \{ L_{N_1}(x,y), T_{N_2}(x,y) \}
\end{equation}
where the range of both $x$ and $y$ coordinates is $[-1,1]$.
Each basis function in $H_{N_1,N_2}(x,y)$ is mapped linearly to the range centered at $[0.9, 1.1]$ for even fairness. We use $H_{3,5}(x,y)$ as basis set.
The neural network $\theta$ predicts the diffused noise in $x_t$ referred to \cite{DDPM,DDIM}, and the loss function is:
\begin{equation}
	L_{BFC}(\theta)= \mathbb{E}\Vert {N_\theta}\left({x_t},t\right)-{N} \Vert_2^2
\end{equation}

\subsection{Metal Artifact Reduction in CT}
\label{E3}
We use the Adam with the momentum as $(0.9,0.999)$. The initial learning rate is set to $1\times 10^{-5}$, decays to $1e^{-8}$ with factor is $0.6$ and  a patience of $25000$ steps. We use $0.9999$ Exponential Moving Average (EMA). GDMP is trained for $500$ epochs and the batch size is $1$.
The total diffusion steps $T$ are set to $100$. The noise schedule is defined as $\bar{\alpha}=\prod_{s=1}^t \alpha_s$ as in \cite{DDPM,DDIM}, where $\alpha_t=1-\beta_t$ and $\beta_t$ is linearly increasing from $0.0001$ to $0.02$.

The metal-affected CT is $X_{ma}$ and the metal-free CT $x_0$. The diffused noise is image difference between $X_{ma}$ and $x_{0}$, the $\eta$ is $10$, denoted by $N=(10+\epsilon)(X_{ma}-x_0)/11,\epsilon\sim \mathcal{N}(0,1)$.
In metal artifact reduction, the metal domain has far more severe artefacts than the non-metal domain, but the restoration of the metal domains is non-essential  and it reduces the prediction accuracy of the artefacts in the non-metal domains. Hence, we optimised the loss function using a weighted MSE that balances the artefact intensity between metal and non-metal domains:
\begin{equation}
	\begin{split}
		L_{MAR}(\theta) =&  \mathbb{E} \left(1+(1-m)\dfrac{max(m\cdot {N})}{max((1-m)\cdot {N})}\right) \notag\\
        & \times \Vert{N_\theta}(x_t,t)-{N}\Vert_2^2
	\end{split}
\end{equation}
where the $m$ is the metal mask.

\subsection{Shadow Removal in Natural Image}
\label{E4}
For shadow removal, the training hyperparameters are consistent with the official implementations in ShadowFormer. 
The total diffusion steps $T$ are set to $100$. The noise schedule is defined as $\bar{\alpha}=\prod_{s=1}^t \alpha_s$ as in \cite{DDPM,DDIM}, where $\alpha_t=1-\beta_t$ and $\beta_t$ is linearly increasing from $0.0001$ to $0.02$.
Since the GDMP adds shadow maps of different intensities during diffusion, keep the coefficients of the shadow mask in the ShadowFormer consistent with the coefficients of the shadow maps, which is $\sqrt{1-\alpha_t}$. The $\eta$ is $10$.

The diffused noise ${N}$ as $(10+\epsilon)({X_S}-{x_0})/11$ similar to MAR, where the ${X_S}$ is shadow image and ${x_0}$ is non-shadow image. 
Our learning objective should be consistent with the baseline model ShadowFormer that directly learns the non-shadow image ${x_0}$. Hence the loss function is:
\begin{equation}
	L_{SR}(\theta)= \mathbb{E} \Vert {x_\theta}({x_t},t)-{x_0} \Vert_2^2
\end{equation}


\end{document}